\documentclass[11pt, a4paper, logo, copyright]{googledeepmind}

\pdftrailerid{redacted}

\makeatletter
\providecommand\bibentry[1]{\nocite{#1}\fullcite{#1}}
\makeatother

\usepackage{kantlipsum, lipsum}
\usepackage{dsfont}
\usepackage{gdm-colors}
\usepackage[utf8]{inputenc}   
\usepackage{newunicodechar}   
\usepackage{amssymb}          
\usepackage{arydshln}
\usepackage{wasysym,marvosym}

\usepackage{xcolor}
\usepackage{soul}

\soulregister\cite7
\soulregister\ref7
\soulregister\cref7

\newcommand{\hlyellow}[1]{{\sethlcolor{yellow!10}\hl{#1}}}
\newcommand{\hlpurple}[1]{{\sethlcolor{purple!10}\hl{#1}}}
\newcommand{\hlblue}[1]{{\sethlcolor{blue!10}\hl{#1}}}

\usepackage[
  style=authoryear-comp,
  sorting=ynt,
  natbib=true,
  backend=bibtex,
  maxcitenames=2,
  maxbibnames=99,
  uniquelist=false,
  uniquename=false,
  giveninits=true,
  dashed=false
]{biblatex}
\let\cite\textcite

\DeclareFieldFormat{citehyperref}{%
  \DeclareFieldAlias{bibhyperref}{noformat}%
  \bibhyperref{#1}}
\DeclareFieldFormat{textcitehyperref}{%
  \DeclareFieldAlias{bibhyperref}{noformat}%
  \bibhyperref{%
    #1%
    \ifbool{cbx:parens}
      {\bibcloseparen\global\boolfalse{cbx:parens}}
      {}}}
\savebibmacro{cite}
\savebibmacro{textcite}
\renewbibmacro*{cite}{%
  \printtext[citehyperref]{%
    \restorebibmacro{cite}%
    \usebibmacro{cite}}}
\renewbibmacro*{textcite}{%
  \ifboolexpr{
    ( not test {\iffieldundef{prenote}} and
      test {\ifnumequal{\value{citecount}}{1}} )
    or
    ( not test {\iffieldundef{postnote}} and
      test {\ifnumequal{\value{citecount}}{\value{citetotal}}} )
  }
    {\DeclareFieldAlias{textcitehyperref}{noformat}}
    {}%
  \printtext[textcitehyperref]{%
    \restorebibmacro{textcite}%
    \usebibmacro{textcite}}}

\usepackage{bbding}
\usepackage[utf8]{inputenc} 
\usepackage[T1]{fontenc}    
\usepackage{hyperref}       
\usepackage{url}            
\usepackage{booktabs}       
\usepackage{nicefrac}       
\usepackage{microtype}      
\usepackage{amsmath}
\usepackage{graphicx}
\usepackage{tablefootnote}
\usepackage{multicol}
\usepackage[nameinlink]{cleveref}
\usepackage{bbm}
\usepackage{multirow}
\usepackage{soul}
\usepackage{float}
\usepackage{wrapfig}
\usepackage{blindtext}
\usepackage{tablefootnote}
\usepackage{amsfonts}
\usepackage[flushleft]{threeparttable}
\usepackage{colortbl}
\usepackage{mathtools,amssymb}
\usepackage{bm}
\usepackage{makecell}
\usepackage{caption}
\usepackage{capt-of}
\usepackage{array}
\usepackage{calc}      
\usepackage{caption}   
\usepackage{subcaption}  
\usepackage{xcolor,colortbl}
\usepackage[bottom]{footmisc}

\usepackage{tcolorbox}

\definecolor{mydeepgreen}{RGB}{0, 100, 0}

\usepackage{xspace}

\newcommand{\myorange}[1]{\textcolor{dmorange500}{#1}}

\newcommand{\cdashlinerow}[2]{%
  \cdashline{#1}%
  \noalign{\global\let\CT@row@color\relax\vskip0pt}%
  \rowcolor{#2}%
}

\newcommand{\eat}[1]{}

\crefformat{section}{\S#2#1#3}

\graphicspath{{figures/}}

\title{Kwai Summary Attention Technical Report}

\renewcommand{\today}{}

\author{\large OneRec Team}

\begin{abstract}

Long-context ability, has become one of the most important iteration direction of next-generation Large Language Models, particularly in semantic understanding/reasoning, code agentic intelligence and recommendation system.
However, the standard softmax attention exhibits quadratic time complexity with respect to sequence length.
As the sequence length increases, this incurs substantial overhead in long-context settings, leading the training and inference costs of extremely long sequences deteriorate rapidly.
Existing solutions mitigate this issue through two technique routings: 
i) Reducing the KV cache per layer, such as from the head-level compression GQA, and the embedding dimension-level compression MLA, but the KV cache remains linearly dependent on the sequence length at a 1:1 ratio.
ii) Interleaving with KV Cache friendly architecture, such as local attention SWA, linear kernel GDN, but often involve trade-offs among KV Cache and long-context modeling effectiveness.
Besides the two technique routings, we argue that there exists an intermediate path not well explored: {Maintaining a linear relationship between the KV cache and sequence length, but performing semantic-level compression through a specific ratio $k$}. This $O(n/k)$ path does not pursue a ``minimum KV cache'', but rather trades acceptable memory costs for complete, referential, and interpretable retention of long distant dependency.
Motivated by this, we propose \textbf{Kwai Summary Attention (KSA)}, a novel attention mechanism that reduces sequence modeling cost by compressing historical contexts into learnable summary tokens.
Specifically, summary tokens are injected into the input sequence at regular intervals, interleaved with text tokens, which partitions the sequence into distinct chunks, and enables text tokens and summary tokens to operate with different attention scopes.
This design ensures both long-range dependencies via summary tokens and full expressiveness via dense attention with text tokens in adjacent chunks.
Empirical results demonstrate that our hybrid-KSA significantly surpass other hybrid variants in long-context modeling, e.g., +3.69\%/5.48\% than hybrid-GDN at RULER-128K at from-scratch/CPT setting.
In inference, KSA's sequence-level compression is fully orthogonal with GQA and MLA. Composing KSA with these methods provides a reliable $8\times$ further KV cache compression while preserving model performance.
\end{abstract}

\begin{document}

\maketitle

\begin{figure*}[th!]
\centering
\includegraphics[width=\textwidth]{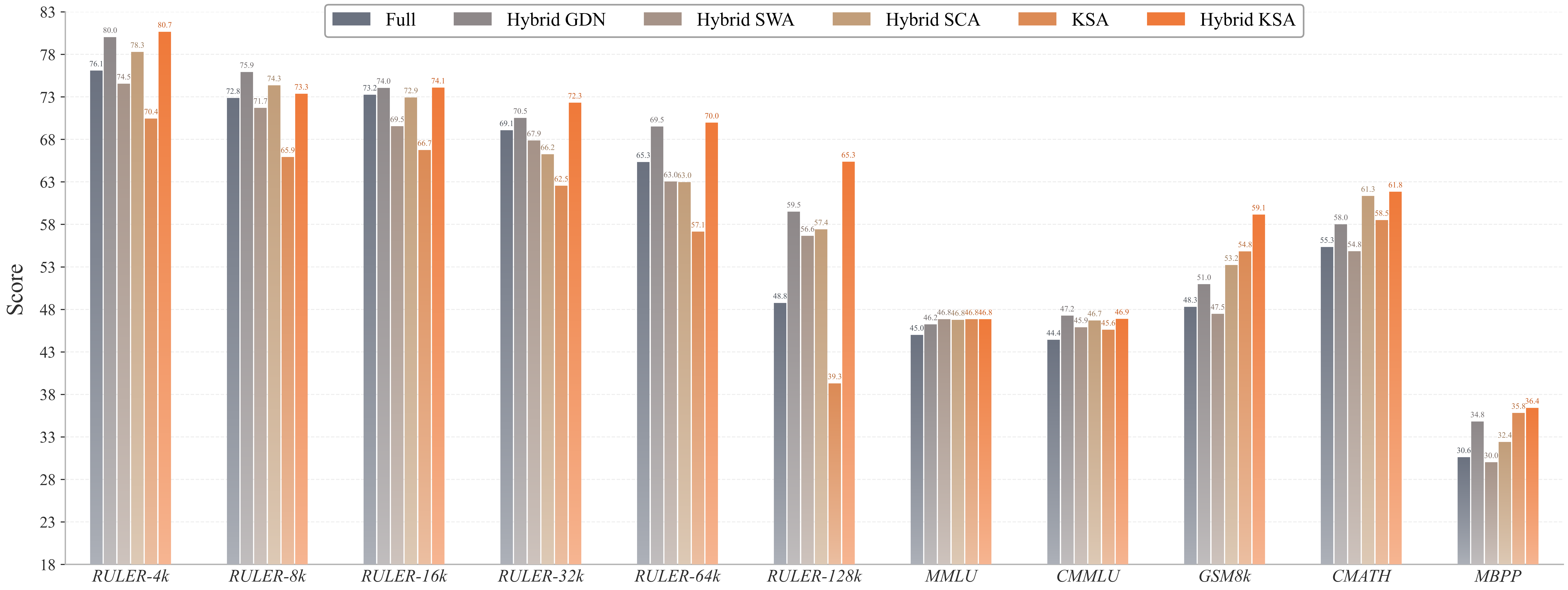}
\caption{From scratch model variants performance on long-context and general benchmarks. Our hybrid-KSA (8$\times$ sequence compression, $3$:$1$ KSA/Full mixture ratio) leads on multiple long-context retrieval tasks and achieves competitive general understanding ability with other Hybrid variants.}
\label{fig:first}
\end{figure*}

\section{Introduction}

In past years, the Transformer's scaling-laws~\cite{kaplan2020scaling} on data volume, model parameters and computation resources achieve great success to compress the knowledge of human society, pushing the AI intelligence frontier again and again.
Excepted the above three well-explored scaling directions, the sequence modeling context length also provides a vital scaling perspective at same time.
With parallel infrastructure improvement, valid context window gradually increases from earliest 1K to the latest 1M sequence~\cite{qwen35blog,team2026longcat,anthropic2026opus,kimi26blog}.
As we know, the long-context modeling capabilities can significantly reduce model hallucination and incubate new Agent applications in solving intricate real-world tasks, such as the Code Agent~\cite{zhang2024codeagent}, OpenClaw~\cite{2026openclaw} and so on.

In the Agentic era, obtaining extensive memory across multi-turn conversations and accurately retrieving historical contexts are prerequisites for highly intelligent agents and long chain-of-thought reasoning.
However, vanilla attention~\cite{vaswani2017attention} is subjected by its Achilles' heel that imposes un-affordable storage and compute overhead as the sequence length grows: the KV cache scales linearly with sequence length, while the attention computation scales quadratically.
To further unleash the long-context capability of LLMs and unlock the next tier of intelligence,  inventing efficient attention variants and designing corresponding modeling architectures that alleviate the computation bottleneck, have become priorities research topic in the LLM field.

Around the challenge of \textit{how to reduce KV cache and computational overhead in long-context setting}, two technique paths have been well explored in recent years:
\begin{itemize}
    \item Reducing the KV cache per layer, evolving from MHA to GQA, \textit{e.g.,} the Qwen series~\cite{yang2024qwen2technicalreport,yang2025qwen3technicalreport,qwen35blog,qwen36plus}, MLA (DeepSeek series~\cite{liu2024mla,liu2024dsv3,guo2025deepseek,deepseekai2026deepseekv4}, Kimi K2~\cite{team2025kimik2}), MLA-DSA (DeepSeek V3.2~\cite{liu2025dsa}, GLM-5~\cite{zeng2026glm} and DeepSeek NSA~\cite{yuan2025native}), as refinement of the naive full attention. 
    The core idea of this branch is to compress the per-token KV cache through head grouping and sharing, or low-rank KV projection.
    However, those efforts' KV cache costs still maintain a strict $1{:}1$ linear relationship with sequence length.
    \item Interleaving with KV Cache friendly architecture, with varying mixing ratios, represented by linear Gated DeltaNet + GQA~\cite{qwen35blog} and SWA + GQA~\cite{agarwal2025gpt,xiao2026mimo}.
    This branch replaces the majority of layers with efficient attention variants that carry a much smaller KV cache, fundamentally decoupling the KV cache from sequence length in some layers.
    However, the drawback is also obvious: for linear attention, the fixed-size state is inherently lossy compression, in which long-range information becomes blurred and unattainable; local variants, on the other hand, completely discard any context outside the window, thus losing perception of the distant information.
\end{itemize}

The above discussion reveals: i) KV cache compression methods, \textit{e.g.,} GQA, MLA still maintain strict linear relationship between sequence length and cache cost. 
ii) Architectures whose KV cache costs are independent of sequence length, \textit{e.g.,} Hybrid SWA, Hybrid Linear, face limited modeling expressivity and impaired long-context capacity. 
However, we argue that there exists a promising intermediate technique roadmap: \textit{Maintaining a linear relationship between the KV cache and sequence length, but performing semantic-level compression through a specific ratio $k$}\footnote{A similar line of sequence-level compression works has explored this direction with learnable summary/gist tokens; we discuss the relations and differences in Sec.~\ref{sec:sequence_level_token_compression}.}.This $O(n/k)$ path does not pursue a ``minimum KV cache'', but rather trades acceptable memory costs for concrete and interpretable long distant dependency.
Compared with SWA and Linear attention, it has the potential that preserving full fidelity over long-range dependency, more friendly for long-context reasoning, agent trajectories, and downstream RL training signals.
\myorange{In other words, we believe that sequence-level token compression could provide a new perspective for reducing KV cache}.
At the same time, DeepSeek V4~\cite{deepseekai2026deepseekv4} model series have been released recently, whose design follows the first-principle of sequence-level KV Cache compression too, further proves the engineering feasibility and long-context robustness of this technique roadmap.

\begin{figure*}[t]
\centering

\begin{subfigure}{\textwidth}
    \centering
    \includegraphics[width=\textwidth]{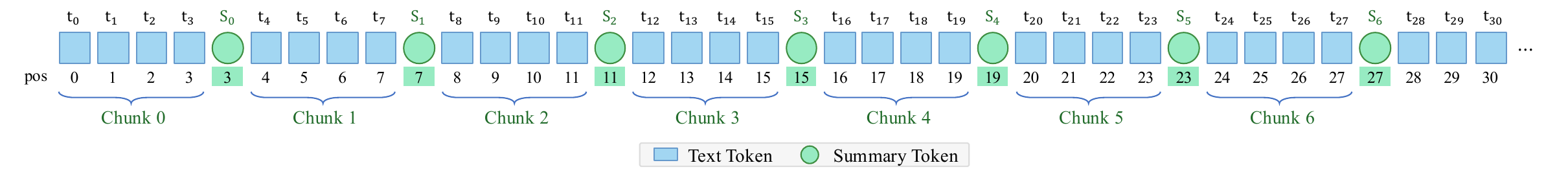}
    \caption{Input sequence is sliced into chunks and summary tokens are injected by the boundary of each chunk.}
    \label{fig:mainmodel_a}
\end{subfigure}

\begin{subfigure}{\textwidth}
    \centering
    \includegraphics[width=\textwidth]{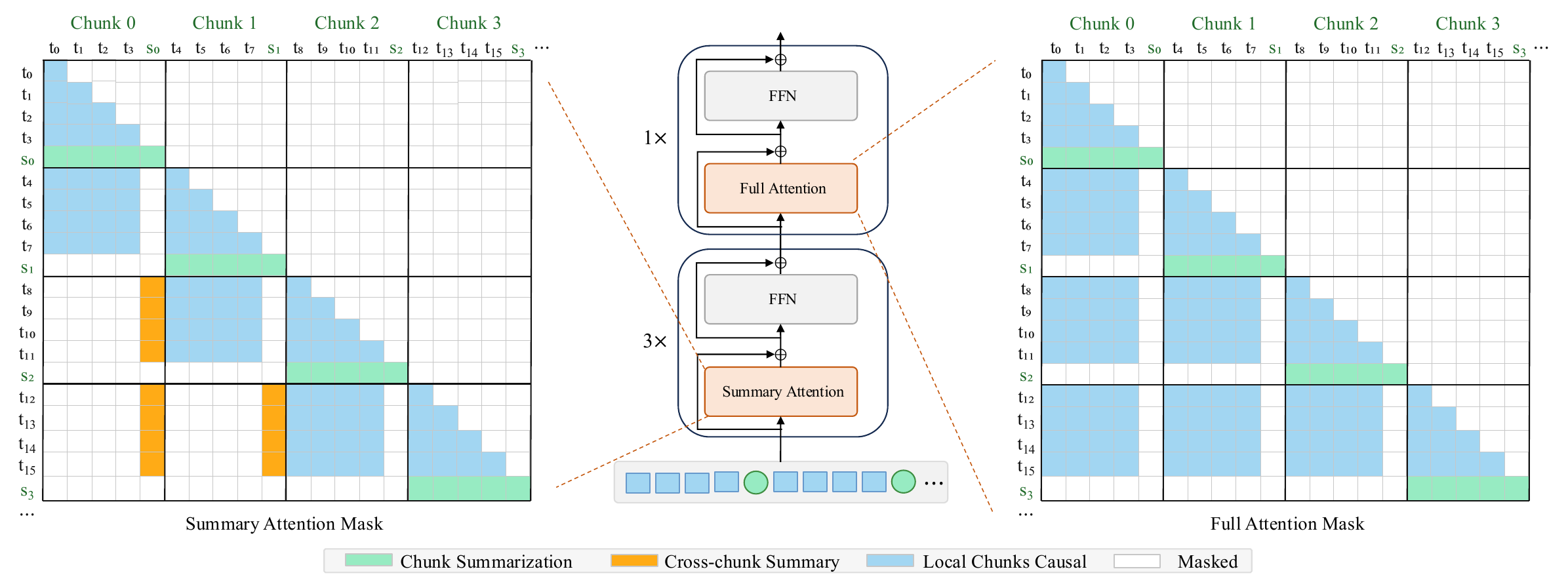}
    \caption{Summary tokens only see within chunk, while text tokens consider both local texts and distant summaries.}
    \label{fig:mainmodel_b}
\end{subfigure}

\caption{Our proposed \textbf{Kwai Summary Attention (KSA)}, maintains sub-quadratic complexity while preserving the expressivity in modeling long-range dependencies. By interleaving KSA with full attention blocks, hybrid-KSA achieves satisfactory balance between performance and efficiency.}
\label{fig:mainmodel}
\end{figure*}

In light of the above, we propose \textbf{Kwai Summary Attention (KSA)}, an efficient, adaptive, and scalable attention mechanism.
KSA introduces a novel summary mechanism designed to distill historical contexts from lengthy sequences into a lightweight, learnable summary token. 
As illustrated in Figure~\ref{fig:mainmodel_a}, our method operates by slicing the input sequence into multiple chunks of a specified size and injecting a learnable summary token at the end of each chunk.
To achieve satisfactory trade-off between modeling capacity and efficiency in long-sequence processing, KSA enables text tokens and summary tokens to perform distinct computational patterns, as shown in Figure~\ref{fig:mainmodel_b}:
i) On the one hand, the summary tokens serve the purpose of distilling semantic information in the current chunk, thus considering only text tokens within the same chunk.
As the sequence grows, these tokens act as concise priors of distant contexts for subsequent text tokens.
ii) On the other hand, for the semantic fluency and stability, text tokens could interact with local neighboring text tokens via a sliding chunk mechanism, and could attend the long-range summary tokens of previous chunks.
The local-text/global-summary visible tokens are able to provide favorable expressiveness and significantly reducing inference latency and KV-cache memory usage.
To validate the effectiveness of KSA, we perform comprehensive from-scratch (Scratch) or continual pre-training (CPT) experiments on a wide range of downstream benchmarks and carry out detailed ablation studies with a series of architectural variants. 
Moreover, we open-source the KSA training recipes and computation kernel to facilitate the next generation LLM architecture research\footnote{Our KSA training scripts at \url{https://github.com/Kuaishou-OneRec/KSA}}.

\section{Methodology}

\subsection{Rethinking Long-Context Modeling}

LLM long-context training and inference faces two main challenges: KV cache growth and attention computation cost.
In retrospect, Full attention and its variants GQA/MLA preserves the complete history, but has KV cache grows linearly with sequence length, which becomes the primary bottleneck for long-context inference. 
In contrast, pure linear or local attention attains linear scaling through fixed-size recurrent states or window size, yet the limited state capacity often fails to retain fine-grained semantic information callback over long contexts effectively.

KSA strikes a trade-off between these two extremes: it continuously compresses long-context information into a growing set of summary states, enabling expressive modeling of distant dependencies without explicitly caching all historical tokens. 
Unlike pure linear attention, the summary state is not fixed-size but grows progressively as summary tokens; unlike sparse attention, long-range modeling no longer hinges on sparse token-to-token connections but routes distant information through summary tokens that act as compressed relays.

Taken together, KSA can be regarded as a mixture mechanism that integrates local-global attention (exact local token-level modeling within a sliding window) with compressed global long-range attention (via a linearly growing set of summary tokens through a specific ratio), offering a better balance among modeling capacity, computational efficiency, and memory footprint.

\subsection{Kwai Summary Attention}
\label{sec:KSA}
This section presents the two key components of Kwai Summary Attention: \textbf{summary token compression} and \textbf{sliding chunk attention}.

\subsubsection{Summary Token Compression}

Given an input sequence $\mathcal{T} = [t_0, \dots, t_{n-1}]$ of $n$ text tokens and a chunk size $k$, we first partition it into $n/k$ chunks (assume $n$ is divisible by $k$) and append a shared learnable summary embedding $s$ at the end of each chunk; we write $s_j$ for the occurrence of $s$ placed in chunk $j$.
Each summary token serves as a distilled representation of text tokens within the chunk.
Formally, the augmented sequence $\hat{\mathcal{T}}$ is:
\begin{align}
\label{eq:summary_compression}
\hat{\mathcal{T}} = [\text{chunk}_{0}, \text{chunk}_{1}, \dots, \text{chunk}_{\frac{n}{k} - 1}], \ \ \text{where} \ \ \text{chunk}_j = [t_{jk}, t_{jk+1},\dots, t_{jk + (k - 1)}, s_j];
\end{align}
where $t_{jk}$ is the first text token of chunk $j$, and $s_j$ is its summary token (all $s_j$ share the special learnable summary token $s$).

Based on the two different token role, we impose structural constraints on the information flow to make the visibility of summary tokens and text tokens complementary:
\begin{itemize}
    \item Summary tokens can only see text tokens within their own chunk, and nothing else. Moreover, summary token position id is the same with its own chunk's last text token's.
    \item Text tokens can see its short-term sliding neighbor text tokens and past long-term summary tokens, without direct access to the full text history.
\end{itemize}
This design explicitly decouples token space from state space: \textit{summary tokens focus on short sequence semantic compression}, while \textit{text tokens access long-range information through previous summary tokens}.

\begin{figure*}[t]
\centering
\includegraphics[width=\textwidth]{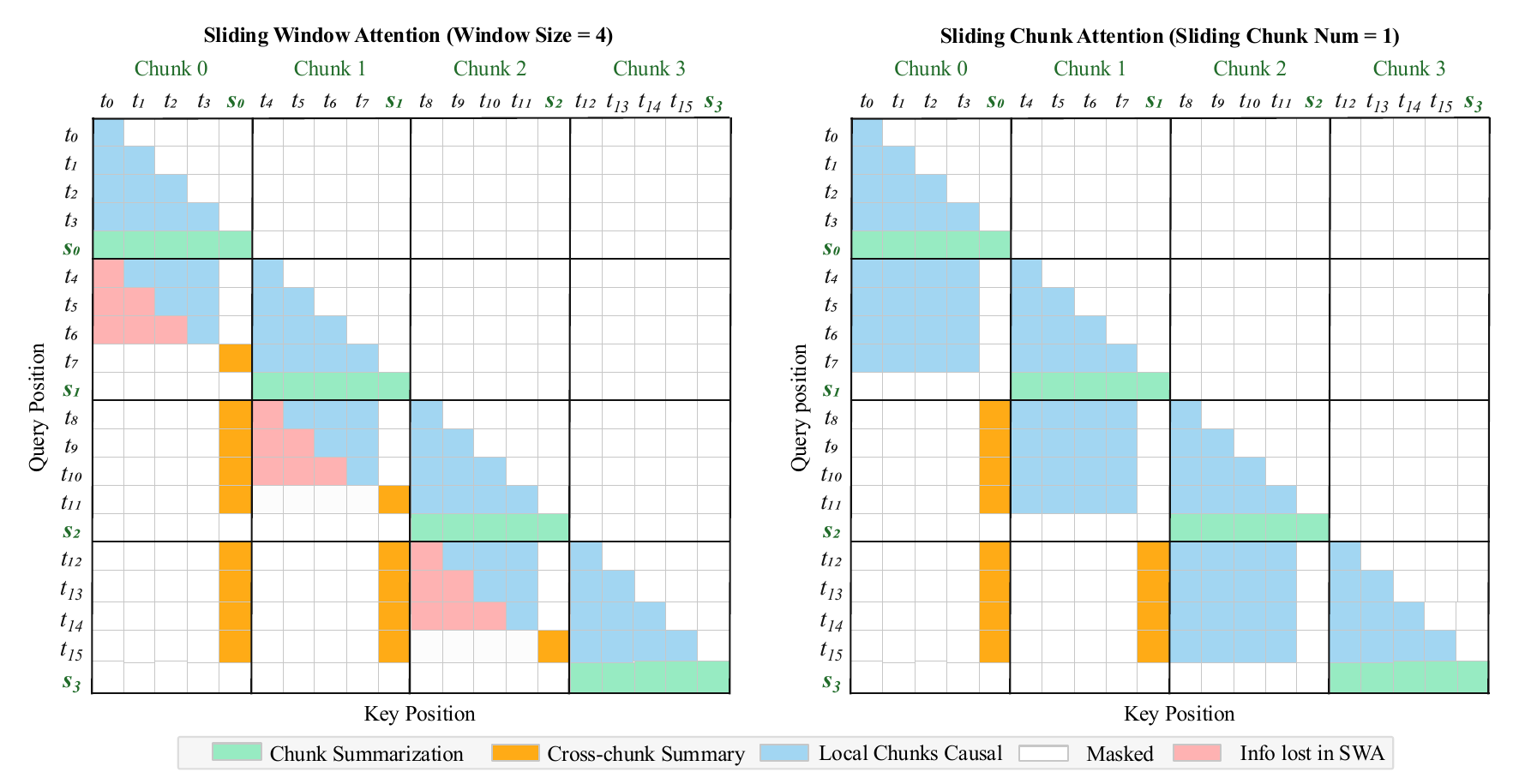}
\caption{\textbf{Sliding Window Attention (left)} may cut through a chunk, resulting attention to overlook certain tokens at boundaries and causing information loss (red regions). \textbf{Sliding Chunk Attention (right)} aligns window boundaries with chunk boundaries, ensuring clean information routing.}
\label{fig:sca_vs_swa}
\end{figure*}

\subsubsection{Sliding Chunk Attention}
\label{sec:sliding_chunk}

To ensure the text token information is orthogonal with the past summary tokens', we further introduce a sliding chunk attention mechanism (SCA) that allows text tokens to see the latest several chunks' text token. 
Formally, for a text token $i$ with chunk size $k$, a standard SCA visible scope is:
\begin{align}
\big[t_{(\lfloor\frac{i}{k}\rfloor - C) * k}, \dots, t_{i-1}, t_i\big]
\end{align}
where $\lfloor\cdot\rfloor$ is the floor division operator, and the $\lfloor\frac{i}{k}\rfloor - C$ indicate the starting chunk index.

In our KSA computation workflow, we combine the SCA and the summary tokens attention together, to inject the short-term neighboring and long-term summary information at the same time.
Formally, for a text token $i$ with chunk size $k$, its text token visible scope as follows:
\begin{align}
\underbrace{\big[s_0, s_1, \dots, s_{\lfloor\frac{i}{k}\rfloor - C -1}\big]}_{\text{distant summary tokens}} \cup \underbrace{\big[t_{(\lfloor\frac{i}{k}\rfloor - C) * k}, \dots, t_{i-1}, t_i\big]}_{\text{sliding chunk text tokens}}
\label{eq:ksa_scope}
\end{align}
The left term covers distant summaries beyond the window; the right term covers text tokens within the window. 
Note that summary tokens within the sliding chunk are \textit{not} considered by text tokens, since their information is already included, as shown in Figure~\ref{fig:sca_vs_swa} attention mask matrix.

\textbf{Why chunk-level sliding, not token-level?} A natural alternative is standard token-level Sliding Window Attention (SWA), where text token could attends to a fix window size latest tokens and it have been already validated in many different LLMs (e.g., GPT-OSS).
However, when combined with our summary tokens, its token-level sliding may omit information or introduce more information. 
As shown in Figure~\ref{fig:sca_vs_swa}, if the window boundary cuts through a chunk, the text token sees only a subset of that chunk's text tokens. 
Meanwhile, the chunk is not fully outside the window, so its summary token does not qualify as a distant summary either. 
As a result, the partially visible chunk is neither covered by complete text nor compensated by its summary, which results in loss of information.

By sliding at chunk granularity, we guarantee a clean partition: every past chunk is either fully within the window (all text tokens visible) or fully outside (accessed only through its summary token), with no intermediate state. This information routing ensures that no chunk's information is lost.

\begin{figure*}[t]
\centering
\begin{subfigure}{\textwidth}
    \centering
    \includegraphics[width=0.9\linewidth]{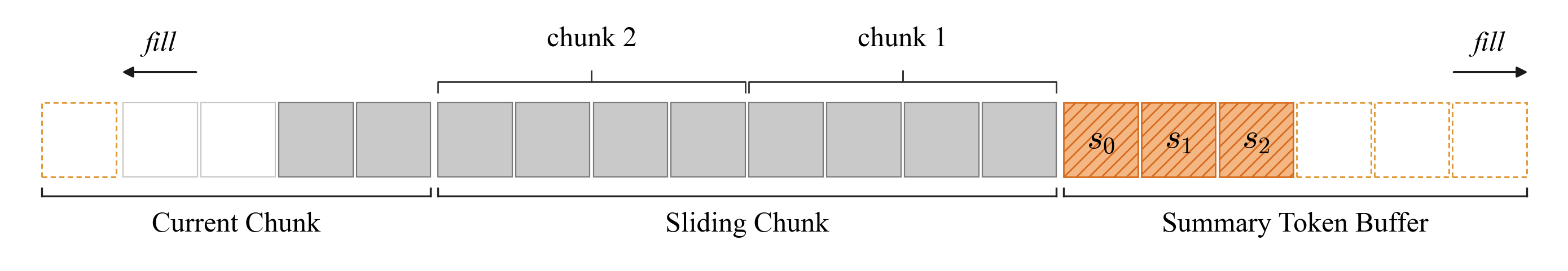}
    \caption{The KSA KV cache consists of contiguous tensors with an organized layout, enabling reading KV states with only one memory operation without concatenation during decoding.}
    \label{fig:kv_cache_a}
\end{subfigure}

\vspace{4pt}

\begin{subfigure}{0.48\textwidth}
    \centering
    \includegraphics[width=\linewidth]{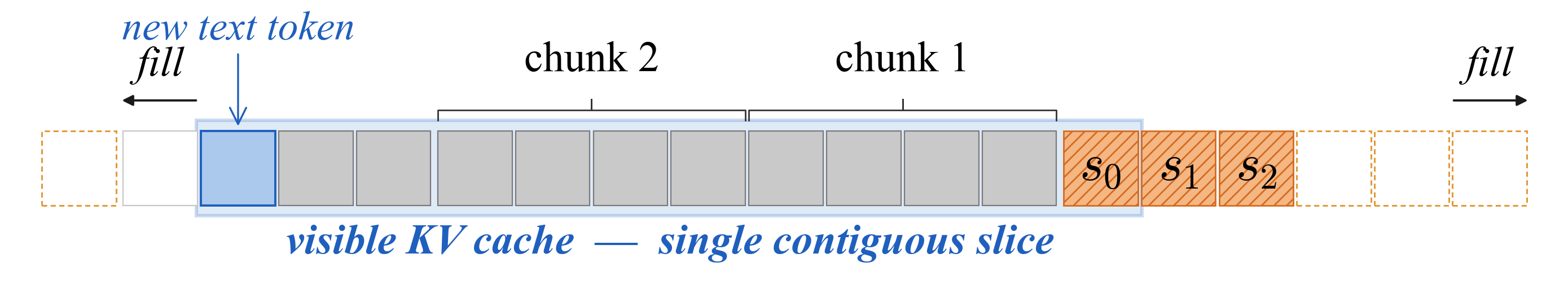}
    \caption{Inserting a new text token: the KV is written into the next available slot of the current chunk, and attention reads a single contiguous slice.}
    \label{fig:kv_cache_b}
\end{subfigure}\hfill
\begin{subfigure}{0.48\textwidth}
    \centering
    \includegraphics[width=\linewidth]{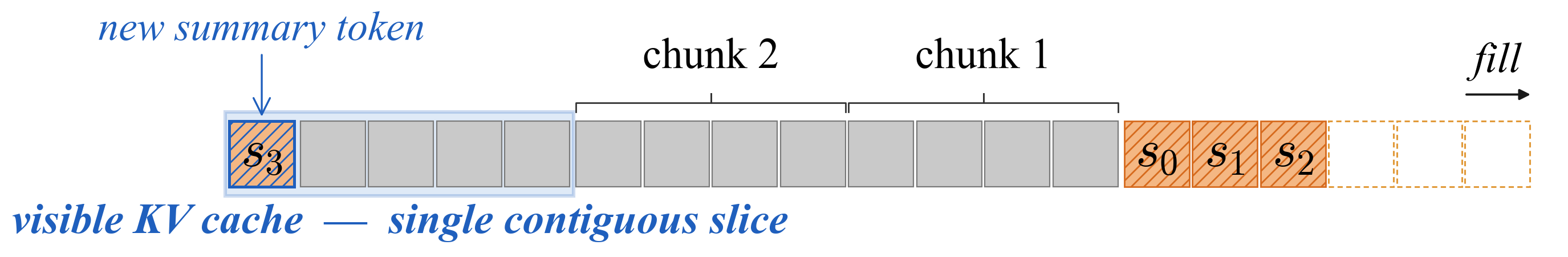}
    \caption{Inserting the chunk summary token: the self-KV is written into the scratch slot left of the current chunk, so summary attention is also a single contiguous slice.}
    \label{fig:kv_cache_c}
\end{subfigure}

\vspace{4pt}

\begin{subfigure}{0.48\textwidth}
    \centering
    \includegraphics[width=\linewidth]{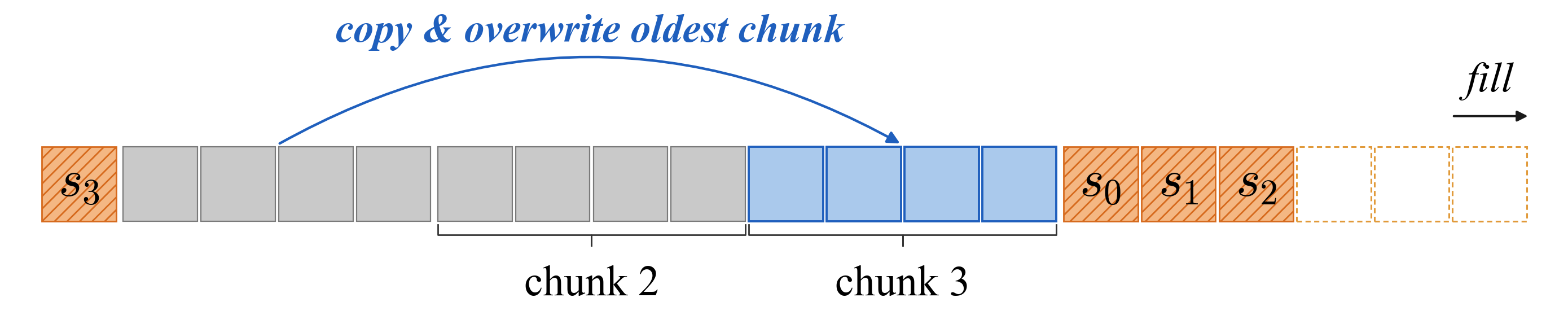}
    \caption{Replacing the oldest text chunk: the just-finished chunk is copied to the ring's write pointer, overwriting the oldest sliding chunk in place.}
    \label{fig:kv_cache_d}
\end{subfigure}\hfill
\begin{subfigure}{0.48\textwidth}
    \centering
    \includegraphics[width=\linewidth]{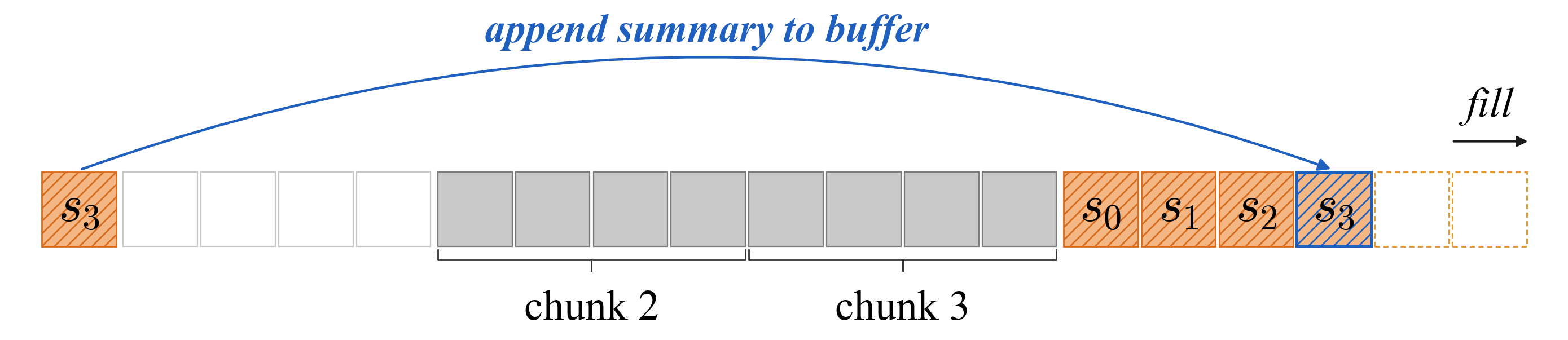}
    \caption{Appending the summary token to the buffer: the new summary token is committed to the right end of the Summary Token Buffer.}
    \label{fig:kv_cache_e}
\end{subfigure}

\caption{The KSA cache features organized layout of contiguous tensors to enable efficient memory access.
The lifecycle involves writing new tokens, inserting summaries, and recycling chunks in place.}
\label{fig:kv_cache}
\end{figure*}

\subsection{Kernel Design}

The attention mask of KSA is a structured sparse mask defined jointly by the local sliding window of text tokens and the visibility of distant summary tokens.
With augmented input of length $L = n + n/k $ where $n/k$ is the total number of summary tokens, building the full mask would cost a prohibitive $O(L^2)$ memory, which is infeasible for long sequences.
To address this, we design a block-sparse attention kernel for training and prefill, and a summary KV cache for decoding, addressing compute efficiency and memory cost respectively.
In the training / prefill kernel, Q/K/V are split into fixed-size blocks, and since KSA discard most Q-K interactions under sparse setting, the kernel only loads non-zero block pairs from HBM to SRAM for computation.

\subsubsection{Efficient Decoding with Summary KV Cache}

The nature of auto-regressive decoding requires every generation step to take all previous KV entries into account, resulting in a memory-bandwidth bottleneck rather than a compute-bound one.
A naive approach requires concatenating, discarding, and re-allocating KV segments at every step, and these scattered memory operations become the performance bottleneck in the tight decode loop.

\paragraph{Composition of KSA KV cache (Figure~\ref{fig:kv_cache_a}).} We organize the KV cache as contiguous tensors, which can be logically divided into three distinct regions: \textit{i) Current Chunk}, \textit{ii) Sliding Chunk Text}, and \textit{iii) Summary Token Buffer}. 
The \textit{Current Chunk} region fills from right to left, with its right edge always aligned to the \textit{Sliding Chunk} region. 
This alignment ensures the caches are physically contiguous, allowing text token attention to read all KV entries in a single operation without concatenation.
The \textit{Summary Token Buffer} grows to the right as new summaries are generated. 
Since RoPE is applied to each key before it enters the cache, every entry inherently carries its own position encoding, ensuring that the physical layout does not interfere with positional information.

\paragraph{Updating the current text chunk (Figure~\ref{fig:kv_cache_b}).} The KV states of incoming tokens are written sequentially into the \textit{Current Chunk} region at the next available slot.
During attention computation, the model retrieves the KV cache as a unified, contiguous slice that spans the current chunk, the sliding chunk text, and the distant summaries.

\paragraph{Chunk eviction and summary token insertion (Figures~\ref{fig:kv_cache_c}, \ref{fig:kv_cache_d}, \ref{fig:kv_cache_e}).} When the current chunk is filled, we execute the following update procedure:
i) The KV of the new summary token is first written into the scratch slot to the left of the \textit{Current Chunk}, so that it stays physically contiguous with the chunk text and summary attention can be computed via a single memory-slicing operation (Figure~\ref{fig:kv_cache_c}).
ii) The just-finished chunk text is then copied into the ring buffer at the write pointer, overwriting the oldest chunk in a modular fashion (Figure~\ref{fig:kv_cache_d}).
iii) Finally, the new summary KV is appended to the \textit{Summary Token Buffer} (Figure~\ref{fig:kv_cache_e}).


Conclusively, every KV read during decoding corresponds to a contiguous slice within the buffer. 
This design eliminates the need for concatenation, gather operations, or explicit mask construction: \textit{the cache layout itself inherently encodes the visibility rule}.

\section{KV Cache Memory Analysis}
\label{sec:kv_cache}

\begin{table*}[t!]
\small
\centering
\caption{Next per-token KV Cache comparison. Taking the $k=8$ chunk compression and $h=128, d=128, g=8, d_c=512, d_r=64$ configurations as example. }
\setlength{\tabcolsep}{5pt}{
\begin{tabular}{lccc}
\toprule
    \textbf{Mechanism} & \textbf{Effective Context} &\textbf{KV Cache Size ($n\rightarrow \infty$)} & \textbf{Compression Rate} \\
    \midrule
    MHA & $n$ (exactly) & $2\cdot n\cdot h\cdot d$  & - \\
    \midrule
    \rowcolor{yellow!10}
    GQA & $n$ (exactly) & $2\cdot n\cdot g\cdot d$ & $g/h \approx$ 6.25\% \\
    \rowcolor{purple!10}
    MLA & $n$ (exactly) & $n \cdot (d_c + d_r)$ & $(d_c + d_r)/(2\cdot h\cdot d) \approx$ 1.76\% \\
    \midrule
    GDN& $n$ (ambiguous) & $2\cdot h \cdot d^2$ & $d/n \approx$ 0\% \\
    \rowcolor{blue!10}
    SWA& $w$ (exactly) & $2\cdot w \cdot g\cdot d$ & $w/n\cdot g/h \approx$ 0\%\\
    \rowcolor{blue!10}
    KSA & $n$ (summarized) & $2\cdot n/k \cdot g\cdot d$ & $1/k \approx$ 12.5\%\\
    \midrule
    KSA with GQA & $n$ (summarized) & $2\cdot n/k \cdot g\cdot d$ & $1/k \cdot g/h \approx$ 0.78\%\\
    KSA with MLA & $n$ (summarized) & $n/k \cdot (d_c + d_r)$ & $1/k \cdot (d_c + d_r)/(2\cdot h\cdot d) \approx$ 0.22\%\\
    \bottomrule
\end{tabular}
}
\label{tab:kv_cache}
\end{table*}

As we known, the key challenge for deploying LLM is the \emph{KV cache state} amount bound during inference.
To alleviate the cache amount issue, there are many comprehensive works were proposed, this section analyzes KSA and other methods KV cache amount in details.

\subsection{Per-Token KV Cache Cost}
Table~\ref{tab:kv_cache} summarizes the KV cache state amount for each mechanism in naive setting:

\paragraph{Naive Multi-Head Full Attention (MHA)~\cite{vaswani2017attention}.} Full Attention stores all $n$ past key-value pairs explicitly, resulting in a cache of size $2nhd$ that grows linearly with sequence length.
Each prediction step requires attending over all $n$ cached KV results.
While this provides access to full context, the linear KV cache growth becomes the notorious bottleneck for long-sequence inference.

\paragraph{Grouped Query Attention (GQA)~\cite{ainslie2023gqa}.} In standard MHA, each of the $h$ attention heads maintains its own KV states, the GQA proposes a simple-yet-effective technique to organize the total $h$ heads into $g$ groups, where same group of Q-heads shared the same cached KV-heads.
Thus the KV cache reduces from $2nhd$ (MHA) to $2ngd$ (GQA), the KV compression ratio is about $g/h$.

\paragraph{Multi-head Latent Attention (MLA)~\cite{liu2024mla}.} Compared with GQA, MLA takes a more aggressive approach by projecting all KV information into low-rank latent vectors. During inference: i) first introduce a $nd_c$ cached block to store the non-RoPE key and values; the per-head non-RoPE keys and values are reconstructed via per-head up-projection matrices. ii) additionally, MLA also introduce a small $nd_r$ cached RoPE block to decoupled for key part. With the two different cache block, the total per-token cache amount is $n(d_c + d_r)$.

\paragraph{Gated Delta Net (GDN)~\cite{yang2025gated}.} Different with the MLA and GQA that follows the standard full attention mechanism, the linear GDN mechanism compresses the entire sequence history into a fixed $d \times d$ state matrix through diagonal gating. The KV cache is always a small constant value $2d^2$, which is independent of sequence length. Nevertheless, the long-term information dependency is unobserved and unexplainable in inference process: as new tokens arrive, older information is progressively attenuated by the decay factors $\alpha_t$, making the effective context is hard to analyze.

\paragraph{Sliding Window Attention (SWA)~\cite{beltagy2020longformer}.} Following the standard full attention mechanism, SWA narrows the visual token window to reduce the KV cache amount. Specifically, the SWA maintains latest $w$ recent tokens in a FIFO buffer, with a fixed cache size $2wgd$. Although effective, the limitation is also obviously: tokens relative index more than $w$ are completely lost, could not access the long-term information directly.

\paragraph{Kwai Summary Attention (KSA)} KSA extends SWA by augmenting the local window with compressed summary tokens. The cache consists of two parts: $2wd$ for the local window (almost same with SWA) and the global $2n/kd$ for the summary tokens, where $k$ is the chunk size, resulting the total cache is about $2wgd + 2(n/k)gd$ and about $2n/kd$ for a large $n$.

\subsection{Orthogonality of KV Cache Compression}
\label{sec:orthogonal}
\begin{figure}[t]
\centering
\includegraphics[width=0.9\columnwidth]{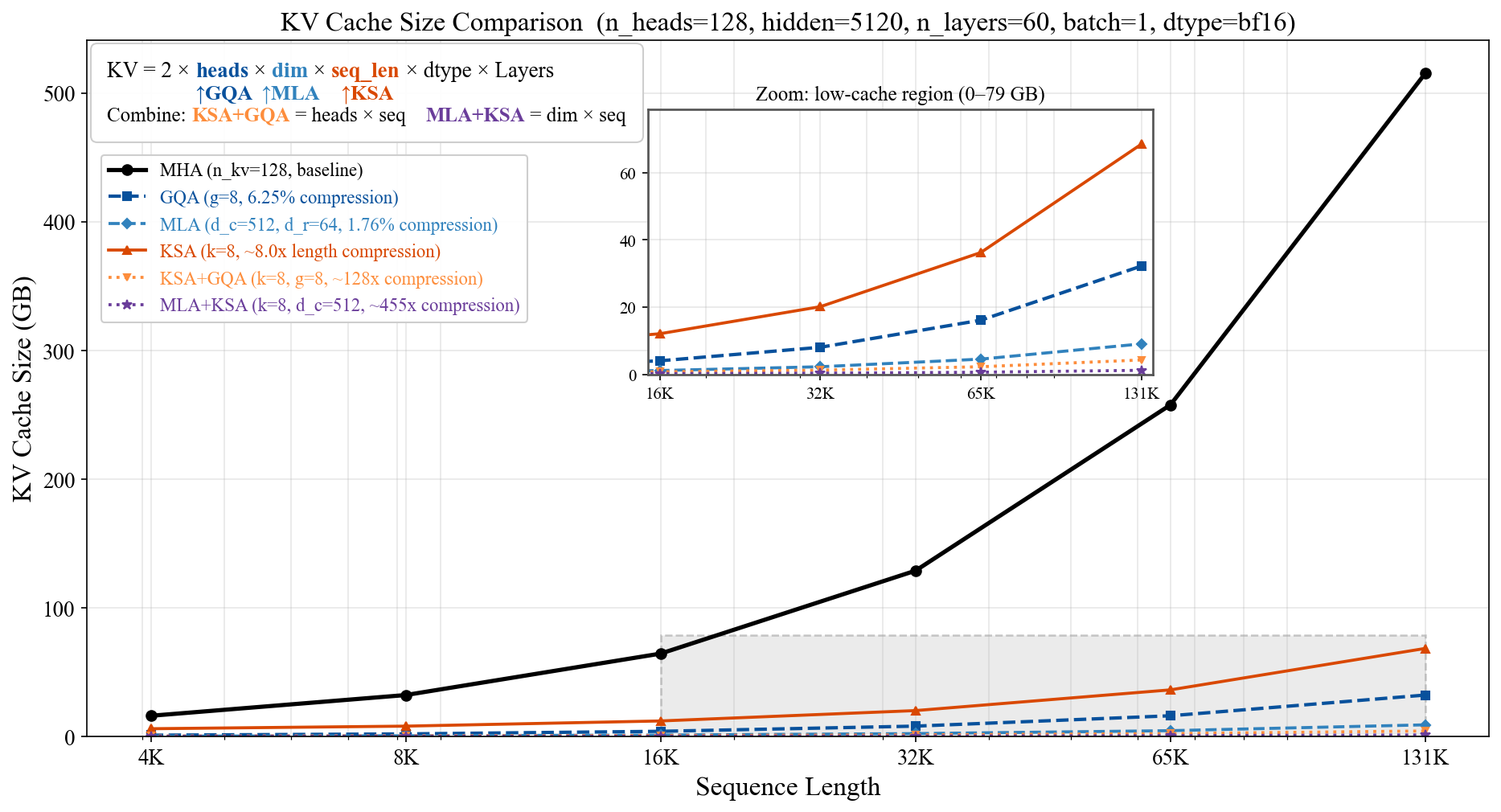}
\caption{KV cache size comparison across different mechanisms as sequence length grows.}
\label{fig:kv_cache_comparison_cross_mechanisms}
\end{figure}

Actually, the KV cache amount of \textbf{one single layer} can be represented as follows:
\begin{equation}
\small
\begin{split}
\text{KV Cache} = \colorbox{blue!10}{\text{PastToken}} \times \colorbox{yellow!10}{\text{HeadNum}} \times \colorbox{purple!10}{\text{EmbeddingDim}}
\end{split}
\label{kvcache}
\end{equation}
Given the above discussion, GQA, MLA, and KSA are orthogonal to each other:
\hlyellow{GQA compresses the heads group},
\hlpurple{MLA shrinks the embedding dimension (which must be re-computed per head)},
and our \hlblue{KSA reduces the number of tokens to be attended}.
Since these three optimization directions are independent,
\textbf{the compression ratios multiply} when combined. For KSA with GQA:

\begin{equation}
  \text{KV Cache}_{\text{KSA+GQA}} = \underbrace{n/k}_{\text{KSA compression}} \times \underbrace{2\cdot g\cdot d}_{\text{GQA compression}}
\end{equation}
In theory, KSA could be combined with MLA:
\begin{equation}
  \text{KV Cache}_{\text{KSA+MLA}} = \underbrace{n/k}_{\text{KSA compression}} \times \underbrace{(d_c + d_r)}_{\text{MLA compression}}
\end{equation}
Table~\ref{tab:kv_cache} summarizes the combined compression for different configurations.
Notably, KSA provides a qualitatively different scaling advantage: while GQA and MLA reduce the constant factor in $O(n)$ cache growth, KSA reduces the growth rate itself to $O(n/k)$. As shown in Figure~\ref{fig:kv_cache_comparison_cross_mechanisms}, for long sequences, this sub-linear scaling could further reach a considerable compression ratio (e.g., 0.78\% in KSA with GQA and 0.22\% in KSA with MLA).

\section{Experiments}

\begin{table}[t]
\centering
\caption{Training model configurations of our KSA.}
\label{tab:model_config}
\small
\renewcommand{\arraystretch}{1.15}
\begin{tabular}{lcc}
\toprule
\textbf{Configuration} & \textbf{From Scratch} & \textbf{Continual Pretraining} \\
\midrule
Number of layers & 24 & 36 \\
Hidden size & 2048 & 2560 \\
Intermediate size & 6144 & 9728 \\
Attention heads (Q/KV) & 16/16 & 32/8 \\
Head dimension & 128 & 128 \\
Hybrid architecture ratio (KSA:Full) & 3:1 & 3:1 \\
Summary chunk size & 8 & 8 \\
Sliding chunk number & 128 & 128 \\
Tied embeddings & False & True \\
\bottomrule
\end{tabular}
\end{table}

\begin{table}[t!]
    \centering
    \small
    \caption{Training hyperparameter settings.}
    \label{tab:train_hyper_param}
    \renewcommand{\arraystretch}{1.2}
    \begin{tabular}{lcc}
    \toprule
    \textbf{Configuration} & \textbf{From Scratch} & \textbf{Continual Pretraining} \\
    \midrule
    Sequence Length Stages & 8K / 32K / 64K / 128K & 32K / 64K / 128K \\
    Token Budget per Stage & 250B / 50B / 50B / 50B & 25B / 35B / 25B \\
    Max Learning Rate 
        & 8K: $4\times10^{-4}$; $\ge$32K: $1\times10^{-5}$ 
        & All stages: $1\times10^{-4}$ \\
    Min Learning Rate & \multicolumn{2}{c}{$1\times10^{-7}$} \\
    RoPE Theta 
        & 8K: $10^4$; $\ge$32K: $10^6$ 
        & $10^6$ \\
    Optimizer & \multicolumn{2}{c}{AdamW ($\beta_1=0.9$, $\beta_2=0.95$)} \\
    Weight Decay & \multicolumn{2}{c}{$0.01$} \\
    LR Schedule & \multicolumn{2}{c}{WSD~\cite{yu2025minicpmv45cookingefficient}} \\
    Gradient Clipping & \multicolumn{2}{c}{$1.0$} \\
    \bottomrule
    \end{tabular}
\end{table}

\subsection{Experiments Setting}

\subsubsection{Model Hyper-parameters Configuration}

We evaluate different attention architectures under two experimental settings: train-from-scratch (Scratch, 400B Token, 128K Sequence) and continual-pretraining (CPT, 85B Token, 128K Sequence), covering both full training and adaptation scenarios.

\paragraph{Baseline models.}
Our tested model variants are as follows: 
(1) \textbf{Pure  Full Attention}, (2) \textbf{hybrid-GDN~\cite{yang2025gated}/Ring-Linear~\cite{team2025every}}, (3) \textbf{Hybrid Sliding Window Attention (SWA)}, (4) \textbf{Hybrid Sliding Chunk Attention (SCA)}, (5) \textbf{Pure KSA}, (6) \textbf{hybrid-KSA}, all of the hybrid model variants with a 3:1 mixing ratio.



\paragraph{Train-from-Scratch.}
We train a 1.9B-parameter language model from scratch using the architecture specified in Table~\ref{tab:model_config} and the training hyperparameters in Table~\ref{tab:train_hyper_param}.
Training follows a progressive length-extension schedule: the model is first trained on 250B tokens at a sequence length of 8K, followed by three subsequent stages of 50B tokens each at 32K, 64K, and 128K, respectively, resulting in a total of 400B tokens across all stages for stable long-context training.

\paragraph{Continual Pretraining.}
For continual pretraining, we initialize from Qwen3-4B-base~\cite{yang2025qwen3technicalreport} with the configuration summarized in Table~\ref{tab:model_config}, and further train on 85B tokens using the hyperparameters in Table~\ref{tab:train_hyper_param}. 
Training is conducted in three stages with progressively increasing sequence lengths, allocating 25B, 35B, and 25B tokens at 32K, 64K, and 128K, respectively. 

\begin{figure*}[t]
\centering
\includegraphics[width=0.99\textwidth]{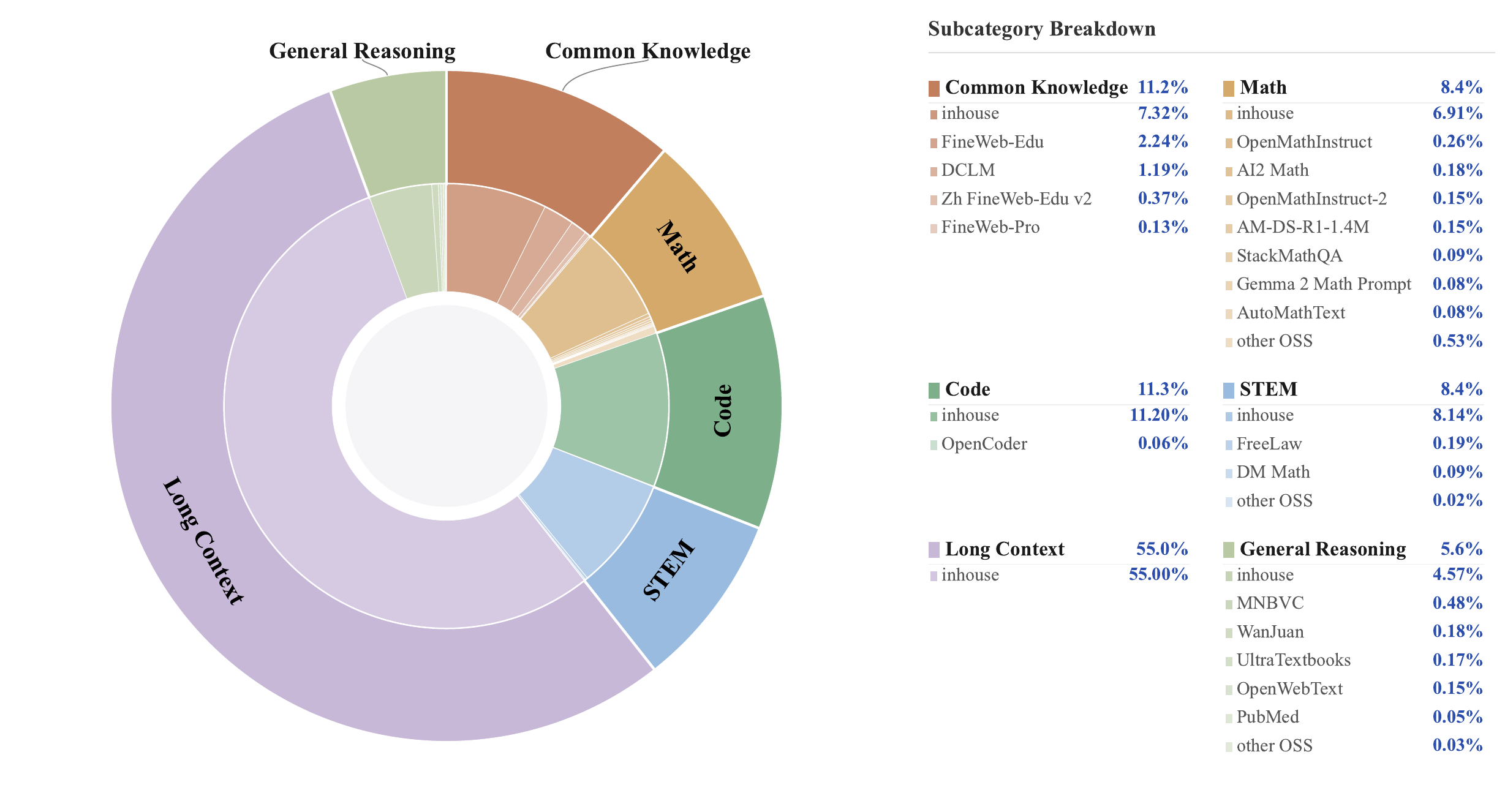}
\caption{Overall Distribution of Pretraining Data Proportions.}
\label{fig:datasets_distribution}
\end{figure*}

\subsubsection{Datasets Pipeline}
The pretraining corpus is composed of six categories: Common Knowledge, Math, Code, STEM, General Reasoning, and Long Context. 
A comprehensive breakdown of data formulation is in Figure~\ref{fig:datasets_distribution}.

\textbf{Long Context} is a primary data class, as it directly serves our goal of reinforcing the model's long-sequence modeling capacity.
We construct a hybrid mixture tailored for context-length extension by combining naturally long documents, \textit{i.e.,} books, long-form QA across code, math, and STEM; 
with synthetic sequences that probe long-range information tracking.
We further incorporate structured benchmark-style sequences from long-context evaluation tasks to ensure robustness.

\textbf{Common Knowledge} forms the foundational layer, sourced from diverse high-quality web text~\cite{penedo2024fineweb}, educational content, and curated open-source corpora. 
The sources span filtered CommonCrawl snapshots~\cite{su2025nemotron}, web-extracted educational material, and high-quality open-domain text, ensuring broad coverage of general world knowledge.

\textbf{Math} and \textbf{Code} are constructed from both natural and synthetic sources.
The math subset integrates web-crawled mathematical content, competition problems, Chinese-language resources, and QA pairs, augmented with synthetic problem, \textit{i.e.,} solution sequences to improve coverage of rare reasoning patterns.
The code subset combines GitHub repositories, competition solutions, code-related QA pairs, and synthetic generation tasks spanning multiple programming languages.

\textbf{STEM} and \textbf{General Reasoning} broaden the model's domain expertise through academic papers, textbook-style content, encyclopedia entries, long-form books, and structured QA corpora.
Domain-specific scientific documents and exam-style data further strengthen coverage of technical disciplines.

The training pipeline follows the data mixture ratios shown in Figure~\ref{fig:datasets_distribution} and progressively extends the context length across stages.
In train-from-scratch experiments, the context length increases from 8K to 32K, 64K, and finally 128K tokens.
In CPT experiments, training starts directly at 32K and advances to 128K, with all stages using the same ratios of data classes mixture.

\subsubsection{Evaluation Benchmark}
We categorize the evaluation benchmarks into two groups, targeting long-context modeling 
ability and general language capability accordingly.

\textbf{Long-Context Benchmarks:} 
RULER~\cite{hsiehruler} is a comprehensive long-context evaluation suite that 
assesses models across multiple dimensions including retrieval, multi-hop tracing, 
aggregation, and question answering, with configurable context lengths up to 128K tokens. 

\textbf{General Capability Benchmarks:} 
We evaluate general capabilities across three domains: general knowledge, mathematics, and code. For general knowledge, MMLU~\cite{hendrycksmeasuring} and its extensions, CMMLU~\cite{li2024cmmlu}, C-Eval~\cite{huang2023c}, and MMLU-Pro~\cite{wang2024mmlu}, assess knowledge breadth and reasoning across diverse domains via multi-choice questions, where CMMLU and C-Eval provide Chinese benchmarks spanning multiple disciplines and exam-style tasks, and MMLU-Pro introduces more challenging, professionally curated questions with reduced data contamination. For mathematics, GSM8K~\cite{cobbe2021training}, CMATH~\cite{wei2023cmath}, and MATH~\cite{hendrycksmath2021} cover problems from grade-school multi-step arithmetic to competition-level symbolic reasoning, including Chinese settings and diverse problem formats requiring multi-step derivations. For code, MBPP~\cite{austin2021program} and HumanEval~\cite{chen2021evaluating} evaluate program synthesis from natural language, with HumanEval further emphasizing functional correctness through unit tests on more complex and realistic coding tasks.

\subsubsection{Training Recipes}
To adapt  summary attention with existing architecture seamlessly, we devise a three stage training stages for CPT: \textit{summary token adaptation}, \textit{parameter annealing}, and \textit{sequence length extension}:

\begin{figure*}[t]
\centering
\includegraphics[width=0.99\textwidth]{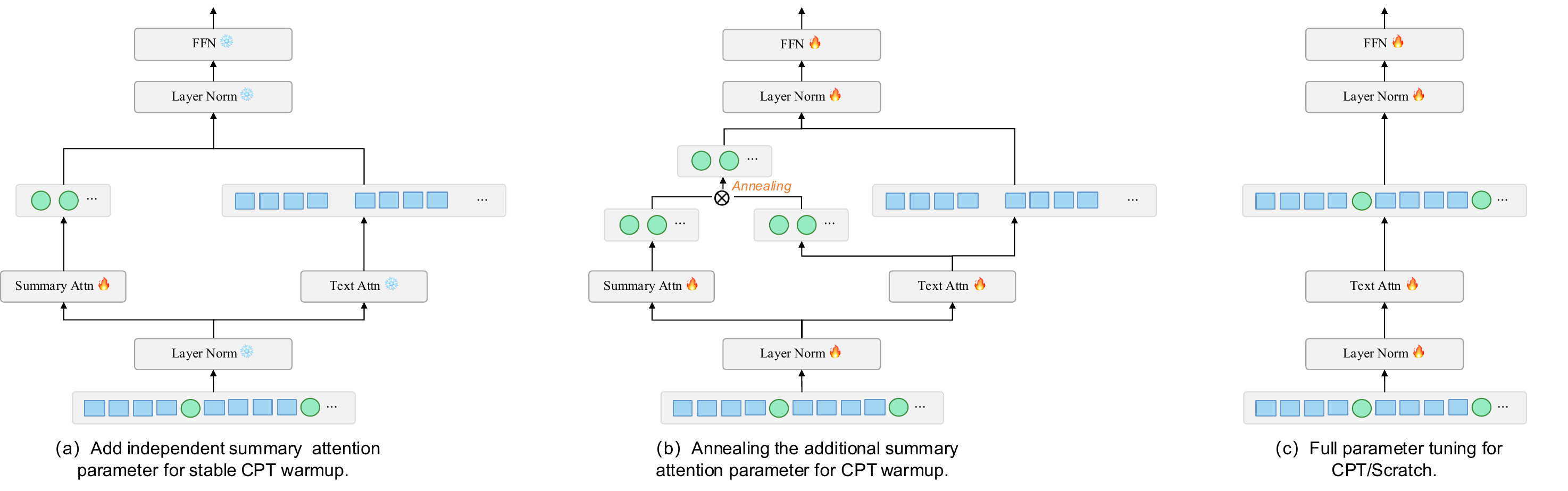}
\caption{The training recipes for CPT warmup: (a) insert additional Q/K/V learnable weights to adapt the summary token compression; (b) conduct the annealing strategy to drop the summary weights; (c) full parameter tuning on large data corpus. Residual connections are omitted for brevity.}
\label{fig:recipes}
\end{figure*}

\paragraph{Summary token adaptation for CPT.}
As described earlier, summary tokens are designed to distill chunk-wise semantic information and serve as compressed priors for distant contexts. 
To endow these tokens with dense, informative representations, we devise a multi-granularity distillation strategy that aligns the KSA student with a vanilla full-attention teacher at three levels: \textit{layer-wise}, \textit{distribution-wise}, and \textit{objective-wise}. 
Concretely, we instantiate the summary token $\mathcal{S}$ as a new vocabulary entry and equip each summary layer with independent attention parameters $W_{\mathcal{S}}^{Q}$, $W_{\mathcal{S}}^{K}$, and $W_{\mathcal{S}}^{V}$.

\textit{Layer-wise attention score alignment.}
Let $W^{Q}, W^{K}, W^{V}$ denote the pretrained attention projections, and let $W^{Q}_{\mathcal{S}}, W^{K}_{\mathcal{S}}, W^{V}_{\mathcal{S}}$ denote the independent summary matrices introduced for KSA. 
Let $\mathcal{T}$ and $\hat{\mathcal{T}}$ denote the original and summary-augmented input sequences (Equation~\ref{eq:summary_compression}), $\mathcal{S} \subset \{1,\dots,|\hat{\mathcal{T}}|\}$ index the summary positions. 
The teacher branch projects input tensors using the pretrained weights,

\begin{equation}
X = \mathcal{T}\,(W^{X})^{\top}, \qquad X \in \{Q, K, V\}.
\end{equation}

The student processes its input by employing position-specific projections: text positions retain the pretrained weights, whereas summary tokens route through the newly introduced parameters,

\begin{equation}
\hat{X}_{t} =
\begin{cases}
\hat{\mathcal{T}}_{t}\,(W^{X})^{\top}, & t \notin \mathcal{S}, \\[2pt]
\hat{\mathcal{T}}_{t}\,(W_{\mathcal{S}}^{X})^{\top}, & t \in \mathcal{S},
\end{cases}
\qquad X \in \{Q, K, V\}.
\label{eq:attn_qkv_for_distill}
\end{equation}

Attention patterns also diverge between them. 
The teacher performs standard full attention:

\begin{equation}
O = \mathrm{softmax}\!\left(QK^{\top} / \sqrt{d}\right) V,
\end{equation}

whereas the student adopts the KSA pattern $\mathcal{M}_{\text{KSA}}$, in which each text token attends to its local sliding-chunk window together with preceding summary tokens, and each summary token attends only to the text chunk it compresses; as detailed in Section~\ref{sec:KSA}:
\begin{equation}
\hat{O} = \mathrm{softmax}\!\left(
   \hat{Q}\hat{K}^{\top} / \sqrt{d} \;+\; \mathcal{M}_{\text{KSA}}
\right) \hat{V}.
\end{equation}
To align the intermediate representations, we discard the summary rows of $\hat{O}$ (which have no counterpart in $\mathcal{T}$), denoted $\hat{O}\big|_{\mathcal{T}} \in \mathbb{R}^{|\mathcal{T}| \times d}$, and employ a mean squared error (MSE) loss:
\begin{equation}
\mathcal{L}_{\text{MSE}}
= \frac{1}{L \cdot |\mathcal{T}|}
  \sum_{\ell=1}^{L} \big\| O_{\ell} - \hat{O}_{\ell}\big|_{\mathcal{T}} \big\|_{2}^{2},
\label{eq:attn_o_for_distill}
\end{equation}
where $L$ is the number of transformer blocks and $|\mathcal{T}|$ is the sequence length.

\textit{Distribution-wise regularization.}
While $\mathcal{L}_{\text{MSE}}$ constrains intermediate layer outputs, it does not directly enforce consistency in the final predictive distributions. 
We therefore introduce a KL regularizer on the output logits to provide a higher-level alignment between the KSA student and the full-attention teacher. 
Let $W_h$ denote the shared LM head, and $h_L$ and $\hat{h}_L$ be the final-layer hidden states of the teacher and student, respectively. 
The two predicted distributions along with KL regularizer are:

\begin{equation}
p = \mathrm{softmax}\!\left(h_L W_h^{\top}\right), \qquad
\hat{p} = \mathrm{softmax}\!\left(\hat{h}_L W_h^{\top}\right), \qquad
\mathcal{L}_{\text{KL}} = \mathrm{KL}(p \,\|\, \hat{p})
= \sum_{v} p_v \log \frac{p_v}{\hat{p}_v}.
\label{eq:kl_for_distill}
\end{equation}

\textit{Objective-wise training.} 
The total distillation objective combines the language modeling loss on the student with the two alignment terms to ensure warmup eventually benefits model performance:

\begin{equation}
\mathcal{L} = \mathcal{L}_{\text{LM}}
           + \alpha\, \mathcal{L}_{\text{MSE}}
           + \beta\, \mathcal{L}_{\text{KL}},
\end{equation}

where $\alpha$ and $\beta$ are hyperparameters tuned on a validation split.

\textit{Analysis.} 
All three loss terms, $\mathcal{L}_{\text{MSE}}$, $\mathcal{L}_{\text{KL}}$, and $\mathcal{L}_{\text{LM}}$; are computed exclusively over text token positions. 
This design ensures dimensional consistency between the teacher and student outputs (as the teacher lacks summary positions) while preserving semantic fidelity. 
Notably, restricting the loss to text positions does not cut off gradient flow to the summary parameters: \textit{the distinctive attention pattern of text tokens, as defined in Equation~(\ref{eq:ksa_scope}), inherently captures the semantics of summary tokens}, enabling gradients to propagate through the attention interactions. 
Through these three complementary, multi-granularity distillation objectives, the independent summary parameters learn to approximate the attention distribution of vanilla full attention as closely as possible, establishing a solid foundation for subsequent full-parameter training and scaling to longer sequences.

\paragraph{Parameter annealing for CPT.}
To avoid introducing additional parameters that would increase inference cost, we propose a parameter annealing strategy that gradually absorbs the independent summary parameters into the main LLM weights. Specifically, at each summary position, we perform \textit{two QKV projections} on the same hidden state: one using the shared LLM weights, gaining
$(q^{\text{main}}_s, k^{\text{main}}_s, v^{\text{main}}_s)$, and one through
the independent summary weights, yielding $(q_s, k_s, v_s)$.
The QKV triplets
fed into the \textit{single attention computation} are then linearly interpolated:
\begin{equation}
\tilde{x}_s = \lambda\, x_s + (1-\lambda)\, x^{\text{main}}_s,
\qquad x \in \{q, k, v\},
\end{equation}
The interpolation coefficient follows an iteration-dependent schedule:
\begin{equation}
\lambda(s) =
\begin{cases}
1, & s \le s_{\text{start}}, \\[2pt]
1 - \dfrac{s - s_{\text{start}}}{s_{\text{end}} - s_{\text{start}}},
 & s_{\text{start}} < s < s_{\text{end}}, \\[6pt]
0, & s \ge s_{\text{end}},
\end{cases}
\end{equation}

 where $s$ denotes the current training step, and $s_{\text{start}}, s_{\text{end}}$ define the annealing window.
When $s\le s_{\text{start}}$, the summary positions are governed entirely by the independent parameters; when $s\ge s_{\text{end}}$, they rely exclusively on the main LLM weights, allowing the auxiliary parameters to be removed at inference time without any architectural modification.
This strategy provides a smooth curriculum that facilitates the transition from a dedicated summary head to a fully shared representation.

\paragraph{Sequence length extension for CPT/Scratch.}
For CPT settings, introduction of summary tokens stops at the context length of 32K, allowing the newly added parameters to learn stable representations.
We then extend the context in two stages: 64K for 35B tokens, followed by 128K for 25B tokens.
This staged schedule enables the summary mechanism to adapt to longer contexts incrementally, equipping KSA with the ability of handling extremely long sequences.
For the Scratch setting, we directly share the attention weights of summary token and text token at the begin, and then training 250B/50B/50B/50B tokens with the sequence length 8K/32K/64K/128K.
Note that we change the RoPE Theta from $10^4$ to $10^6$ after the 8K training finished.

\begin{table}[t!]
    \footnotesize
        \caption{
            For CPT setting, \textbf{hybrid-KSA} achieves \textit{superior long-context retrieval across all RULER lengths}, while both KSA variants closely match or exceed \textbf{Full} attention on knowledge, math, and coding benchmarks, establishing the \textit{smallest performance gap among all sub-quadratic alternatives}.
        }
        \centering
        \small
        \setlength{\tabcolsep}{2pt}{
        \begin{tabular}{lcccccc}
        \toprule
        \multirow{2}{*}{\makecell{Benchmarks}} & \multicolumn{4}{c}{Baselines} & \multicolumn{2}{c}{Ours} \\
        \cmidrule(lr){2-5} \cmidrule(lr){6-7}
         &  \text{Full} & \text{Hybrid-SWA} & \text{Hybrid-SCA} & \text{Hybrid-Linear} & \text{KSA} & \text{Hybrid-KSA} \\
        \midrule
        \rowcolor{gray!10}
        \textbf{Long-Context Retrieval} &   &  &  &  &  &  \\
        RULER-4K            & 92.88 & 91.30 & 86.02 & 86.39 & 91.55 & \textbf{92.97} \\
        RULER-8K            & \textbf{91.38} & 88.03 & 84.28 & 83.86 & 86.78 & 90.53 \\
        RULER-16K           & \textbf{89.12} & 82.87 & 80.67 & 78.06 & 84.78 & 88.86 \\
        RULER-32K           & 84.74 & 78.94 & 76.89 & 76.48 & 80.30 & \textbf{86.65} \\
        RULER-64K           & \textbf{78.16} & 73.88 & 68.88 & 73.50 & 76.09 & 76.04 \\
        RULER-128K          & 65.86 & 66.27 & 60.94 & 67.98 & 66.81 & \textbf{71.67} \\
        \midrule
        \rowcolor{gray!10}
        \textbf{General Knowledge} &  &  &    &  &  &  \\
        MMLU                & \textbf{71.83} & 70.57 & 69.83 & 64.33 & 70.73 & 70.50 \\
        CMMLU               & \textbf{75.00} & 73.69 & 72.59 & 68.41 & 73.29 & 72.63 \\
        C-Eval              & \textbf{73.66} & 72.36 & 71.66 & 67.42 & 72.14 & 72.66 \\
        MMLU-Pro            & \textbf{46.36} & 45.23 & 45.11 & 38.83 & 45.70 & 45.39 \\
        \midrule
        \rowcolor{gray!10}
        \textbf{Mathematics} &  &  &  &    &  &  \\
        CMath               & 83.41 & \textbf{84.84} & 83.16 & 79.09 & 84.58 & 84.25 \\
        GSM8K               & \textbf{82.75} & 81.92 & 80.10 & 72.44 & 81.09 & 79.50 \\
        MATH                & 47.48 & \textbf{48.24} & 47.45 & 42.57 & 48.15 & 47.56 \\
        \midrule
        \rowcolor{gray!10}
        \textbf{Code} &  &  &  &  &    &  \\
        MBPP                & 61.30 & 61.70 & 59.60 & 55.30 & 61.50 & \textbf{62.20} \\
        HumanEval           & 58.54 & 61.89 & 61.89 & 54.58 & 60.97 & \textbf{62.50} \\
        \midrule
        \rowcolor{gray!20}
        \textbf{Average}    & 73.50 & 72.12 & 69.94 & 67.28 & 72.30 & \textbf{73.59} \\
        \bottomrule
        \end{tabular} }
        \label{tab:cpt}
\end{table}

\subsection{Continual Pre-Training Performance Comparison}

We evaluate the effectiveness of KSA under the CPT setting, comparing full \textbf{KSA} and \textbf{Hybrid-KSA} against four representative baselines: \textbf{Full} attention, sliding window attention (\textbf{Hybrid-SWA}), sliding chunk attention (\textbf{Hybrid-SCA}), and linear attention (\textbf{Hybrid-Linear}). 

\subsubsection{Long-Context Benchamrk}

We first examine KSA on RULER~\cite{hsiehruler}, a long-context stress test where it is designed to excel, against the previously introduced baselines.
From Table~\ref{tab:cpt}, we can draw several conclusions:

i) \textit{\textbf{Hybrid-KSA} demonstrates the strongest long-context retrieval capacity among all baselines.}
Specifically, it achieves the best results at \textbf{RULER-4K (92.97)}, \textbf{RULER-32K (86.65)}, and \textbf{RULER-128K (\textbf{71.67})},
outperforming the \textbf{Full} attention baseline while operating with substantially lower cost.

ii) \textit{Summary tokens effectively compress and propagate distant context information when full attention becomes prohibitive.} 
Notably, at the longest 128K context, \textbf{Hybrid-KSA} surpasses \textbf{Full} attention by \textbf{+5.81} points and outperforms the strongest hybrid baseline (\textbf{Hybrid-SWA}) by \textbf{+5.40} points.

iii) \textit{KSA's way of information aggregation is a more faithful approximation of full attention than fixed-window or linearized alternatives.} 
Compared to other hybrid variants (\textbf{SWA}, \textbf{SCA}, \textbf{Linear}), our \textbf{KSA} and \textbf{Hybrid-KSA} models consistently lead by clear margins across all RULER lengths.

\subsubsection{General Benchmarks}

To test if our architectural design and training procedure preserve model's general ability on real-world problems, we further validate KSA on multiple general benchmarks. 
The evaluation covers three knowledge domains: general knowledge (MMLU, CMMLU), mathematics (CMath, GSM8K), and code generation (MBPP). 
From Table~\ref{tab:cpt}, we observe the following:

i) \textit{CPT of summary attention preserves the pretrained model's general world knowledge.}
On MMLU and CMMLU, the full \textbf{KSA} model achieves 70.73 and 73.29, closely matching the \textbf{Full} attention upper bound (71.83 / 75.00), and substantially outperforms \textbf{Hybrid-Linear} (64.33 / 68.41), which suffers from the limited representation capacity of fixed-size memory updates. 

ii) \textit{KSA is capable of mathematical reasoning.}
The \textbf{KSA} model attains 84.58 on CMath, surpassing \textbf{Full} attention (83.41), and remains competitive on GSM8K (81.09 vs.\ 82.75).

iii) \textit{Our method robustly scales to real-world coding problems.}
\textbf{Hybrid-KSA} achieves \textbf{the best MBPP score (62.20) across all configurations}, including \textbf{Full} attention.

iv) \textit{The proposed approaches consistently yield smaller capability gaps relative to \textbf{Full} attention} than other sub-quadratic baselines across all general-purpose benchmarks.

\begin{table}[t!]
    \footnotesize
        \caption{
            \textbf{Hybrid-KSA} consistently outperforms full and sub-quadratic baselines, achieving superior long-context scalability and strong general performance in from-scratch training.
        }
        \centering
        \small
        \setlength{\tabcolsep}{2pt}{
        \begin{tabular}{lcccccc}
        \toprule
        \multirow{2}{*}{\makecell{Benchmarks}} & \multicolumn{4}{c}{Baselines} & \multicolumn{2}{c}{Ours} \\
        \cmidrule(lr){2-5} \cmidrule(lr){6-7}
         &  \text{Full} & \text{Hybrid-SWA} & \text{Hybrid-SCA} & \text{Hybrid-GDN} & \text{KSA} & \text{Hybrid-KSA} \\
        \midrule
        \rowcolor{gray!10}
        \textbf{Long-Context Retrieval} &   &  &  &  &  &  \\
        RULER-4K           & 76.08 & 74.54 & 77.72 & 79.83 & 70.44 & \textbf{80.65} \\
        RULER-8K           & 72.85 & 71.69 & 75.22 & \textbf{76.01} & 65.91 & 73.35 \\
        RULER-16K          & 73.24 & 69.54 & 72.55 & 74.04 & 66.74 & \textbf{74.07} \\
        RULER-32K          & 69.06 & 67.86 & 67.74 & 70.41 & 62.54 & \textbf{72.30} \\
        RULER-64K          & 65.32 & 63.03 & 63.54 & 69.39 & 57.13 & \textbf{69.95} \\
        RULER-128K         & 48.75 & 56.64 & 58.01 & 59.87 & 39.29 & \textbf{65.35} \\
        \midrule
        \rowcolor{gray!10}
        \textbf{General Knowledge} &  &  &    &  &  &  \\
        MMLU               & 44.99 & \textbf{46.84} & 46.77 & 46.23 & 46.83 & 46.83 \\
        CMMLU              & 44.41 & 45.89 & 46.42 & \textbf{47.19} & 45.59 & 46.88 \\
        C-Eval             & 44.28 & 43.54 & \textbf{47.62} & 45.54 & 45.69 & 44.13 \\
        MMLU-Pro           & 19.48 & 20.46 & 20.10 & 21.22 & 21.72 & \textbf{22.52} \\
        \midrule
        \rowcolor{gray!10}
        \textbf{Mathematics} &  &  &  &    &  &  \\
        CMath              & 55.33 & 54.83 & \textbf{62.33} & 58.00 & 58.50 & 61.83 \\
        GSM8K              & 48.29 & 47.46 & 52.39 & 50.95 & 54.81 & \textbf{59.14} \\
        MATH               & 23.38 & 31.46 & 28.82 & 33.30 & 30.04 & \textbf{36.92} \\
        \midrule
        \rowcolor{gray!10}
        \textbf{Code} &  &  &  &  &    &  \\
        MBPP               & 30.60 & 30.00 & 31.60 & 34.80 & 35.80 & \textbf{36.40} \\
        HumanEval          & 25.61 & 28.05 & 26.83 & 27.44 & 29.88 & \textbf{31.71} \\
        \midrule
        \rowcolor{gray!20}
        \textbf{Average}   & 49.44 & 50.12 & 51.84 & 52.95 & 48.73 & \textbf{54.80} \\
        \bottomrule
        \end{tabular} }
        \label{tab:scratch}
\end{table}

\subsection{Train-from-scratch Performance Comparison}

We further evaluate KSA in a train-from-scratch setting, where all modules are optimized without pretrained initialization.
This setting isolates the effect of the open-sourced model capabilities based on massive corpora, providing a stricter test of scalability and learning dynamics.
We compare \textbf{KSA} and \textbf{Hybrid-KSA} against \textbf{Full}, \textbf{Hybrid-SWA}, \textbf{Hybrid-SCA}, and \textbf{Hybrid-GDN}.


\subsubsection{Long-Context Benchmark}

Our methods showcase promising performance on the RULER benchmark as in Table~\ref{tab:scratch}:

i) \textit{\textbf{Hybrid-KSA} achieves the best overall performance, surpassing even \textbf{Full} attention by a large margin.}
It consistently outperforms all baselines across nearly all context lengths, achieving the best scores at \textbf{RULER-4K (80.65)}, \textbf{32K (72.30)}, \textbf{64K (69.95)}, and \textbf{128K (65.35)}. Notably, at \textbf{128K}, it surpasses \textbf{Full} attention by a large margin (\textbf{+16.60}), demonstrating superior scalability.

ii) \textit{KSA significantly improves robustness at extreme context lengths.}
While \textbf{Full} attention and other hybrid baselines degrade rapidly as sequence length increases: \textit{e.g.,} \textbf{Full}: 76.08 $\rightarrow$ 48.75, \textbf{Hybrid-KSA} maintains reliably performing, \textit{i.e.,} 80.65 $\rightarrow$ 65.35, indicating strong robustness for long-inputs.

iii) \textit{Summary-based aggregation provides a stronger alternative to existing efficient attention designs.}
Compared to \textbf{Hybrid-SWA} and \textbf{Hybrid-SCA}, which are constrained by local windows or chunking, and \textbf{Hybrid-GDN}, which relies on compressed memory updates, \textbf{Hybrid-KSA} consistently achieves higher accuracy, especially in the long-context regime, \textit{e.g.,} +5.48 over \textbf{GDN} at 128K.

\subsubsection{General Benchmarks}

We next evaluate general capabilities across knowledge, mathematics, and code benchmarks:

i) \textit{Our method maintains competitive general knowledge performance.}
On MMLU and CMMLU, \textbf{Hybrid-KSA} achieves \textbf{46.83} and \textbf{46.88}, matching or exceeding most baselines and remaining close to the best-performing methods, indicating no loss in general understanding ability.

ii) \textit{KSA demonstrates strong gains in mathematical reasoning.}
\textbf{Hybrid-KSA} achieves the top on GSM8K (\textbf{59.14}) and MATH (\textbf{36.92}), outperforming \textbf{Full} attention by \textbf{+10.85} and \textbf{+13.54}, respectively. This suggests our method can effectively support multi-step reasoning over long contexts.

iii) \textit{Summary attention improves performance on code generation tasks.}
On MBPP and HumanEval, \textbf{Hybrid-KSA} achieves the highest scores (\textbf{36.40} and \textbf{31.71}), outperforming \textbf{Full} attention and all hybrid baselines, demonstrating strong capability in structured generation.

iv) \textit{The proposed method provides a better efficiency–performance trade-off than prior sub-quadratic methods.}
Across all benchmarks, \textbf{Hybrid-KSA} consistently ranks among the top performers, while maintaining sub-quadratic complexity. Compared to other efficient attention mechanisms, it achieves a substantially smaller performance gap, or even improvements, relative to \textbf{Full} attention.

\begin{figure}[t!]
\centering
\captionsetup{justification=centering}
\captionsetup[subfigure]{justification=centering}
\begin{subfigure}{0.9\linewidth}
\centering
\includegraphics[width=\linewidth]{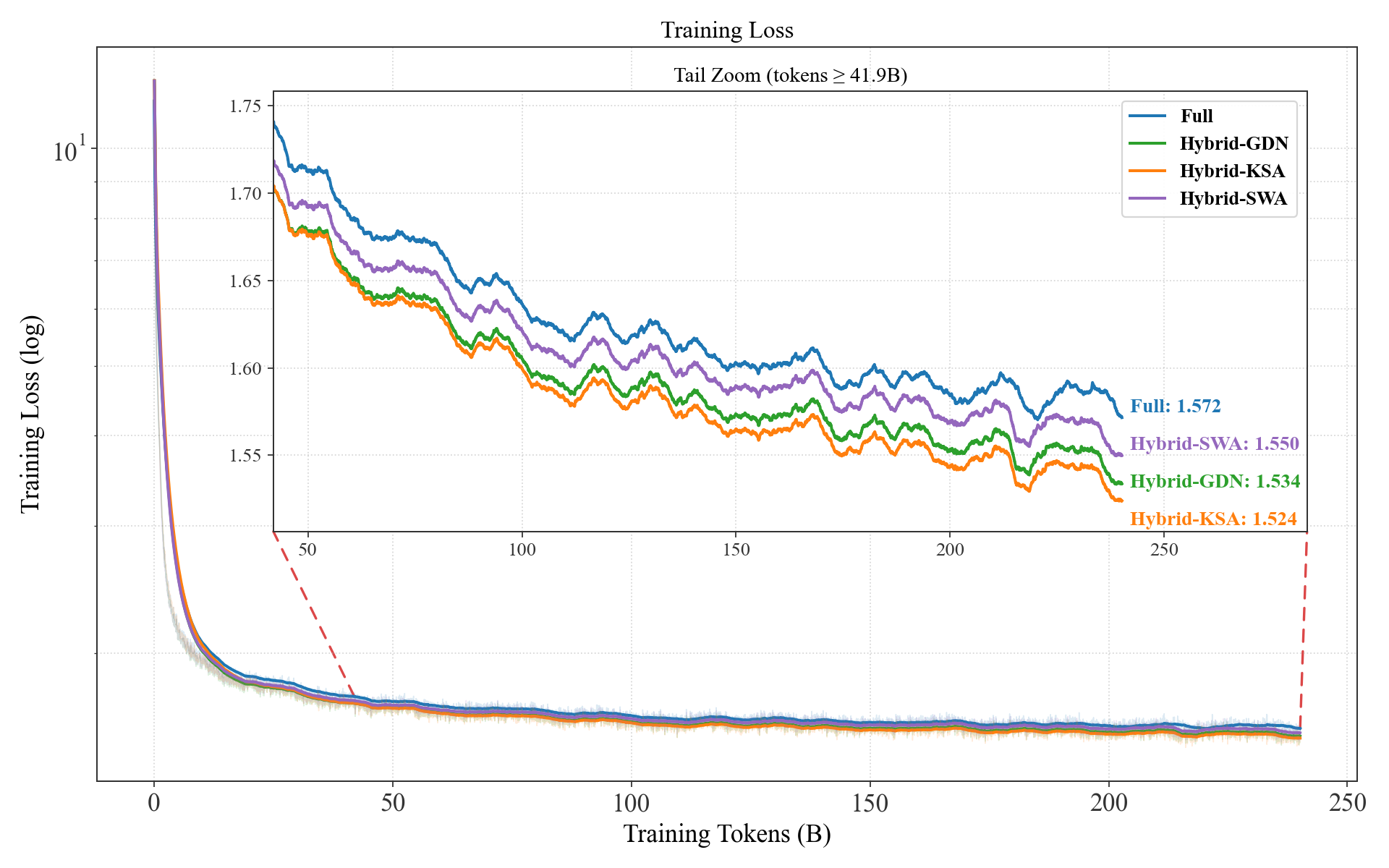}
\caption{\textbf{Hybrid-KSA} has the best convergence efficiency during from-scratch training.}
\label{fig:scratch_loss}
\end{subfigure}
\vspace{0.5em}
\begin{subfigure}{\linewidth}
\centering
\includegraphics[width=\linewidth]{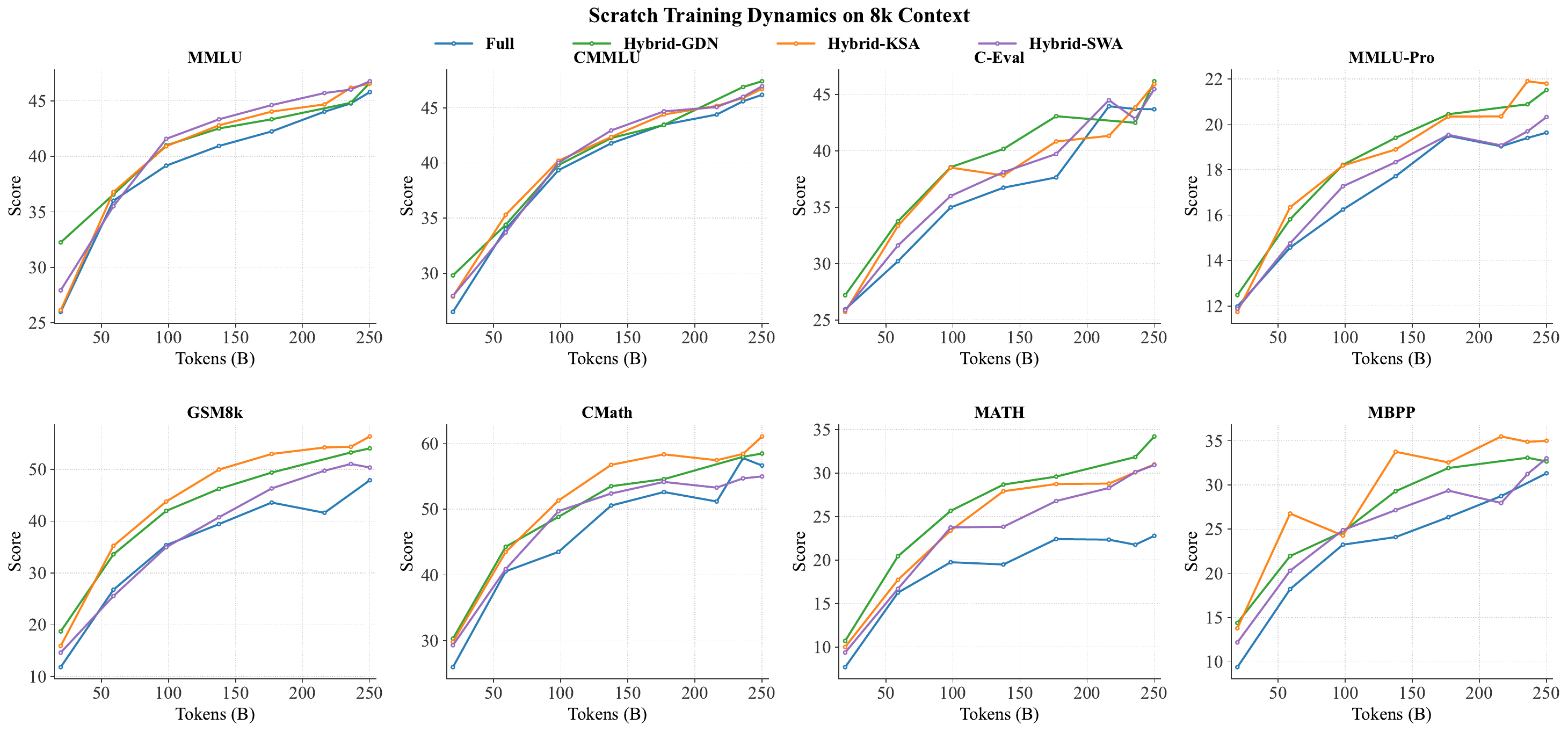}
\caption{Our method's performance improves more rapidly than other models on multiple benchmarks.}
\label{fig:scratch_8k_trends}
\end{subfigure}
\caption{\textbf{Hybrid-KSA} delivers the best training dynamics, with superior convergence efficiency and faster performance gains across benchmarks in from-scratch training.}
\label{fig:scratch_8k}
\end{figure}

\subsubsection{Training Loss and Evaluation Score Analysis}

We further analyze the training dynamics of Hybrid-KSA by examining both training loss and evaluation performance across multiple benchmarks under the train-from-scratch setting.

\paragraph{Training Loss.}
\textbf{Hybrid-KSA} demonstrates best converge efficiency compared to all baselines:

i) \textit{Hybrid-KSA achieves the lowest training loss throughout training.}
As shown in Figure~\ref{fig:scratch_loss}, at the end of training, \textbf{Hybrid-KSA} reaches the lowest loss (\textbf{1.524}), outperforming \textbf{Hybrid-GDN} (1.534), \textbf{Hybrid-SWA} (1.550), and \textbf{Full} attention (1.572).

ii) \textit{The advantage becomes more pronounced in the long-tail regime.}
In the zoomed-in region (tokens $\geq$ 41.9B), Hybrid-KSA consistently stays below all baselines with a stable margin.

\paragraph{Evaluation Scores.}
The improved optimization translates directly into stronger performance:

i) \textit{Hybrid-KSA consistently matches or outperforms baselines across benchmarks.}
Across knowledge (MMLU, CMMLU, C-Eval), reasoning (GSM8K, MATH), and code (MBPP) tasks, Hybrid-KSA demonstrates competitive or superior performance throughout training. Notably, it achieves the best final scores on several tasks, including \textbf{GSM8K}, \textbf{MATH}, and \textbf{MBPP}.

ii) \textit{Hybrid-KSA exhibits faster and more stable performance improvement.}
Compared to \textbf{Full} attention, which lags behind across most benchmarks, Hybrid-KSA shows consistently higher scores at intermediate training stages, suggesting improved learning efficiency. It also avoids the fluctuations observed in other hybrid baselines (e.g., \textbf{Hybrid-SWA}), leading to smoother and more reliable gains.

iii) \textit{The advantage is particularly evident in reasoning-intensive tasks.}
On GSM8K, CMath, and MATH, Hybrid-KSA maintains a clear margin over \textbf{Full} attention and remains competitive with or better than other hybrid methods, indicating that the summary-based mechanism does not hinder, and may even enhance, multi-step reasoning ability.

\begin{table}[t!]
    \footnotesize
        \caption{
            For CPT, \textbf{Hybrid-KSA} achieves the best overall results on RULER 128K subtasks, beating all other efficient baselines and \textbf{Full} attention on \textbf{NIAH-Multivalue}, \textbf{VT}, \textbf{FWE}, and \textbf{SQuAD}.
        }
        \centering
        \small
        \setlength{\tabcolsep}{3pt}{
        \begin{tabular}{lcccccc}
        \toprule
        \multirow{2}{*}{\makecell{Subtasks}} & \multicolumn{4}{c}{Baselines} & \multicolumn{2}{c}{Ours} \\
        \cmidrule(lr){2-5} \cmidrule(lr){6-7}
         & \text{Full} & \text{Hybrid-SWA} & \text{Hybrid-SCA} & \text{Hybrid-Linear} & \text{KSA} & \text{Hybrid-KSA} \\
        \midrule
        NIAH-Single     & 100.00 & 100.00 & 99.16  & 100.00 & 97.50  & 100.00 \\
        NIAH-Multikey   & 75.00  & 74.16  & 70.84  & \textbf{79.16} & 74.16 & 75.84 \\
        NIAH-Multivalue & 88.12  & 83.75  & 91.25  & 95.62  & 83.75  & \textbf{98.75} \\
        NIAH-Multiquery & 95.62  & 93.12  & 98.12  & \textbf{99.38} & 95.62 & 98.12 \\
        VT              & 60.50  & 67.50  & 42.50  & 87.50  & 65.50  & \textbf{90.50} \\
        FWE             & 51.66  & 51.66  & 33.33  & 23.33  & \textbf{72.50} & 65.84 \\
        SQuAD           & 30.00  & 30.00  & 15.00  & 35.00  & 32.50  & \textbf{42.50} \\
        \bottomrule
        \end{tabular} }
        \label{tab:ruler_subtask}
\end{table}

\begin{figure}[t!]
\centering
\includegraphics[width=0.9\linewidth]{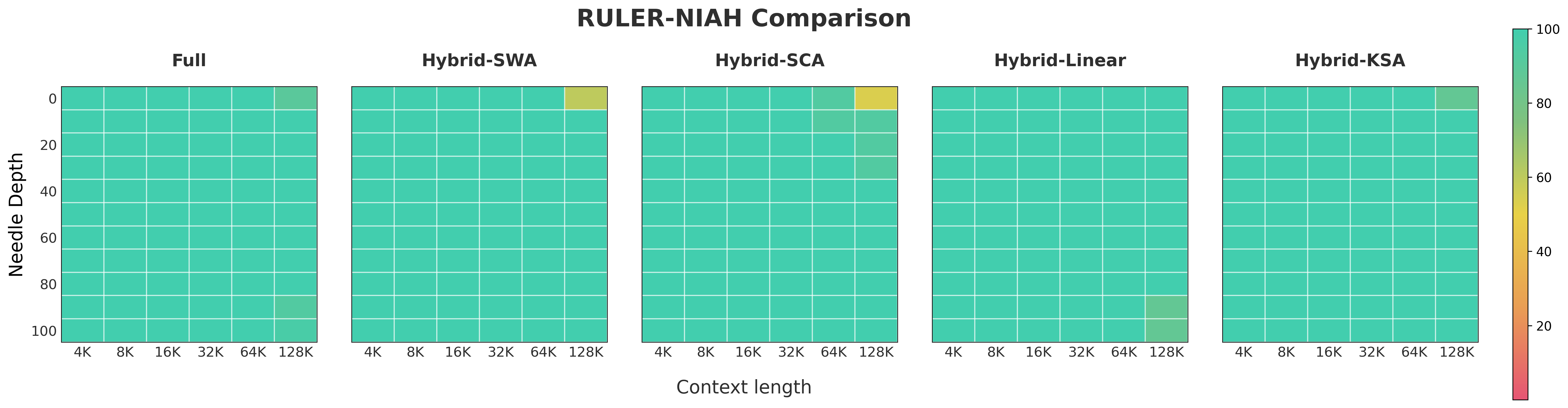}
\caption{NIAH-Single results after converting the model into a hybrid architecture and applying continual pre-training. \textbf{Hybrid-KSA} achieves near-perfect needle retrieval up to 128K context length, with only a minor drop from full accuracy at 128K.}
\label{fig:needle-in-a-haystack}
\end{figure}

\subsection{Needle-in-a-Haystack}

To delve deeper into KSA's ability to retrieve specific information from long contexts, we conduct the Needle-in-a-Haystack benchmark~\cite{martin2023NIAH}.
In this task, a short factual statement, \textit{i.e.,} the \textit{needle}, is embedded at varying depths within a long context, \textit{i.e.,} the \textit{haystack}; and the model is asked to retrieve it via a targeted question.
The retrieval success rate across different needle positions serves as a robust indicator of long-context information retrieval capability.

Figure~\ref{fig:needle-in-a-haystack} reports single-needle retrieval accuracy across context lengths from 4K to 128K and needle depths from 0\% to 100\%.
\textbf{Hybrid-KSA} attention maintains near-perfect retrieval across all lengths and depths, with only a slight dip at 128K, suggesting that a small dose of full attention effectively compensates for the compression loss in extremely long sequences.

Further quantitative analysis in Table~\ref{tab:ruler_subtask} reveals that \textbf{Hybrid-KSA} consistently outperforms other methods with limited effective context across multiple subtasks:

i) \textit{KSA achieves strong NIAH performance, even compared with \textbf{Full} attention.}
On NIAH-Multivalue, \textbf{Hybrid-KSA} attains \textbf{98.75}, a substantial gain of \textbf{+10.63} over \textbf{Full} attention (88.12) and the \textbf{highest} across all configurations.
On NIAH-Multiquery, it scores 98.12, closely approaching the best result (99.38 by \textbf{Hybrid-Linear}) and surpassing \textbf{Full} attention (95.62).

ii) \textit{Our method scales robustly to more challenging synthetic subtasks beyond simple retrieval.}
\textbf{Hybrid-KSA} surpasses \textbf{+30.0} over \textbf{Full} on VT and \textbf{+14.18} over \textbf{Full} on FWE.

iii) In summary, \textit{KSA serves as a high-fidelity compressed relay, enabling robust information retrieval and complex long-sequence reasoning across 128K tokens without the prohibitive overhead of full attention.}

\subsection{Kwai Summary Attention Design Ablation Analyses}

\subsubsection{Inference KV Cache and Speed Analysis}

\begin{figure}[t]
\centering
\includegraphics[width=1.0\columnwidth]{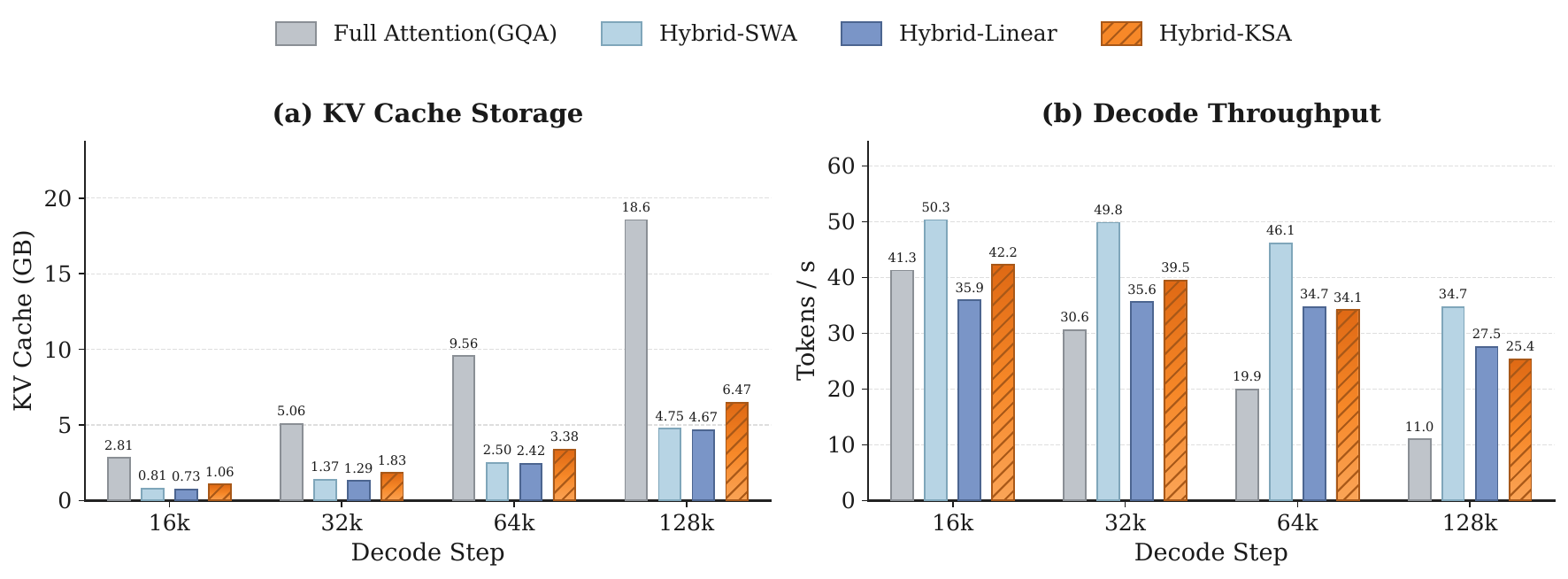}
\caption{Hybrid-KSA reduces KV cache by up to 2.5× while maintaining competitive or higher decode throughput compared with other efficient baselines.}
\label{fig:kv_cache_comparison}
\end{figure}

To provide an intuitive understanding of KSA's efficiency, we test the KV cache memory footprint across context lengths 16-128K and the decode throughput at 16K, and plot the results in Figure~\ref{fig:kv_cache_comparison}:

i) \textit{\textbf{Hybrid-KSA} retains a substantially smaller cache footprint than \textbf{Full} attention}
At 128K, it consumes only 6.47\,GB, which is \textbf{2.8$\times$ smaller} than \textbf{Full} attention (18.6\,GB).

ii)  \textit{Our method achieves higher decode throughput than other efficient baselines.} 
At 128K, \textbf{Hybrid-KSA} achieves a throughput of \textbf{2.3$\times$} relative to \textbf{Full} attention, surpassing both \textbf{Hybrid-SWA}  and \textbf{Hybrid-Ring-Linear} which are 0.73$\times$ and 0.81$\times$ of \textbf{Full} attention respectively.

iii) These results demonstrate that KSA offers a favorable trade-off: \textit{it compresses long-range context into a compact state, reclaiming memory without sacrificing decoding speed}.

\subsubsection{Hybrid-KSA Configuration Analysis}

We conduct ablations on key design choices of Hybrid-KSA, including chunk number N, chunk size S, and the hybrid ratio between summary and full attention; and put the results in Table~\ref{tab:ksa-ablation}.

\paragraph{Chunk Number (N).}
Varying N controls the granularity of summarization. 
As shown in Table~\ref{tab:ksa-ablation}, N=128 strikes a balance between stable long-context ability and competitive general performance:

i) \textit{Moderately increasing N improves long-context performance, but larger values yield diminishing or even negative returns.}
The RULER average improves from \textbf{82.80} to \textbf{83.16} as N increases from 32 to 64. However, further increasing N does not bring consistent gains and can degrade performance at longer ranges: \textit{e.g.,} RULER-64k: 76.35 at N=128 vs.\ 65.73 at N=256.

ii) \textit{N has limited impact on general benchmarks.}
Performance remains largely stable across configurations, with slight improvements at larger N: \textit{e.g.,} reasoning avg: 75.89 at N=256).

\paragraph{Chunk Size.}
Chunk size determines the local receptive field within each segment. We observe a clear trade-off between long-context modeling and general capability:

i) \textit{Smaller chunks yield the stronger long-context performance.}
For example, S=8 achieves the best RULER average (\textbf{82.97}) and 64K performance (\textbf{76.35}).

ii) \textit{Larger chunks improve general performance at the cost of long-context retrieval.}
For instance, S=32 attains the best reasoning average (\textbf{75.23}) and strong GSM8K performance (\textbf{81.23}), while its RULER average drops to 81.50.

iii) \textit{The default choice S=8 provides a balanced anchor point, slightly favoring long-context modeling while retaining strong general capability.}

\paragraph{Hybrid Ratio.}
Hybrid ratio is crucial for model to leverage different architectures' strengths:

i) \textit{Increasing the proportion of summary attention improves long-context performance but weakens general capabilities.}
For example, moving from 3:1 to 5:1 improves the RULER average (\textbf{82.97} $\rightarrow$ \textbf{83.84}), but degrades reasoning (74.48 $\rightarrow$ 74.15) and code performance (MBPP: 60.20 $\rightarrow$ 59.20).

ii) \textit{Increasing full attention enhances general performance at the cost of long-context retrieval.}
For instance, a 1:1 ratio improves reasoning (Cmath: \textbf{84.50}) but significantly reduces long-context performance (RULER avg: 78.72).

iii) \textit{The default choice 3:1 provides a balanced trade-off, achieving strong long-context performance while maintaining competitive general capability.}

\begin{table*}[t!]
    \footnotesize
    \centering
    \caption{The default setting achieves the best balance between long-context and general performance.}
    \label{tab:ksa-ablation}
    \setlength{\tabcolsep}{3pt}
    \begin{tabular}{l cccccc ccc cccc}
    \toprule
    \multirow{2}{*}{Configuration} & \multicolumn{6}{c}{Long-Context (RULER)} & \multicolumn{3}{c}{Knowledge} & \multicolumn{4}{c}{Reasoning \& Code} \\
    \cmidrule(lr){2-7} \cmidrule(lr){8-10} \cmidrule(lr){11-14}
    & 4K & 8K & 16K & 32K & 64K & \textbf{Avg.} & MMLU & CMMLU & \textbf{Avg.} & GSM8K & CMath & MBPP & \textbf{Avg.} \\
    \midrule
    \rowcolor{gray!10}
    \multicolumn{14}{l}{\textit{(a) Chunk Num ($N$)}} \\
    $N=32$ & 91.02 & \textbf{89.46} & \underline{84.42} & \underline{78.35} & \underline{70.74} & 82.80 & 69.93 & \textbf{72.68} & \underline{71.30} & \textbf{81.23} & \underline{83.59} & 60.50 & \underline{75.11} \\
    $N=64$ & \textbf{93.36} & \underline{88.65} & \textbf{87.21} & 76.91 & 69.69 & \textbf{83.16} & \textbf{70.43} & 72.27 & \textbf{71.35} & 80.13 & 83.00 & \underline{61.20} & 74.78 \\
    $N=128$ (Default) & 88.69 & 88.01 & 83.62 & 78.19 & \textbf{76.35} & \underline{82.97} & \underline{70.18} & 72.16 & 71.17 & 80.75 & 82.48 & 60.20 & 74.48 \\
    $N=256$ & \underline{92.05} & 88.30 & 83.86 & \textbf{79.40} & 65.73 & 81.87 & 69.94 & \underline{72.53} & 71.23 & \underline{81.16} & \textbf{84.92} & \textbf{61.60} & \textbf{75.89} \\
    \midrule
    \rowcolor{gray!10}
    \multicolumn{14}{l}{\textit{(b) Chunk Size ($S$)}} \\
    $S=8$ (Default) & 88.69 & \textbf{88.01} & \textbf{83.62} & \textbf{78.19} & \textbf{76.35} & \textbf{82.97} & \textbf{70.18} & 72.16 & \textbf{71.17} & 80.75 & 82.48 & 60.20 & 74.48 \\
    $S=16$ & \underline{88.77} & 83.78 & \underline{82.20} & \underline{77.34} & 71.97 & 80.82 & 69.75 & \textbf{72.50} & \underline{71.13} & \underline{80.81} & \textbf{83.59} & \underline{61.20} & \underline{75.20} \\
    $S=32$ & \textbf{90.21} & \underline{84.66} & 80.26 & 77.31 & \underline{75.09} & \underline{81.50} & \underline{69.91} & 71.99 & 70.95 & \textbf{81.23} & \underline{83.25} & \underline{61.20} & \textbf{75.23} \\
    $S=64$ & 86.50 & 81.61 & 78.12 & 72.63 & 70.09 & 77.79 & 69.83 & \underline{72.28} & 71.05 & 80.48 & 81.84 & \textbf{62.10} & 74.80 \\
    \midrule
    \rowcolor{gray!10}
    \multicolumn{14}{l}{\textit{(c) Hybrid Ratio (Summary : Full)}} \\
    $1{:}1$ & 84.50 & 82.85 & 82.16 & 74.76 & 69.36 & 78.72 & 69.39 & \textbf{72.61} & 71.00 & \textbf{81.09} & \textbf{84.50} & \textbf{60.40} & \textbf{75.33} \\
    $3{:}1$ (Default) & 88.69 & \underline{88.01} & \underline{83.62} & \underline{78.19} & \textbf{76.35} & \underline{82.97} & \textbf{70.18} & 72.16 & \textbf{71.17} & \underline{80.75} & 82.48 & \underline{60.20} & \underline{74.48} \\
    $5{:}1$ & \textbf{93.75} & \textbf{89.12} & \textbf{87.54} & \textbf{78.47} & \underline{70.31} & \textbf{83.84} & \underline{69.47} & \underline{72.55} & \underline{71.01} & 79.83 & \underline{83.41} & 59.20 & 74.15 \\
    $8{:}1$ & \underline{90.80} & 85.88 & 78.94 & 71.83 & 66.19 & 78.73 & 68.31 & 71.06 & 69.69 & 79.98 & 82.91 & 58.10 & 73.67 \\
    \bottomrule
    \end{tabular}
\end{table*}


\subsubsection{Per-Layer Attention Pattern Analysis}
\label{sec:per_layer_attention_pattern_analysis}

\begin{figure}[t]
\centering
\includegraphics[width=\textwidth]{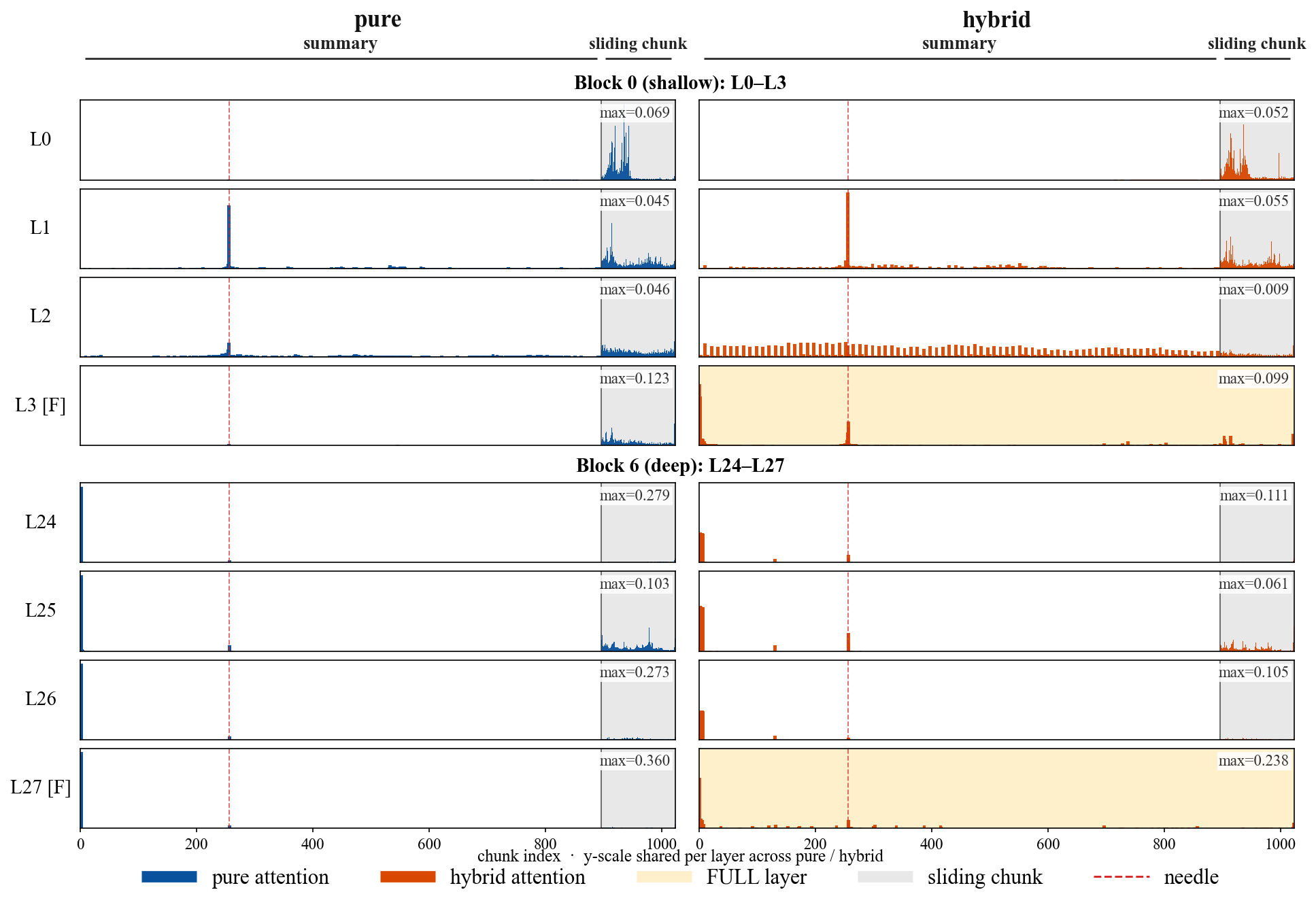}
\caption{Per-layer attention patterns reveal that \textbf{Hybrid-KSA} enhances cross-chunk retrieval through ``comb-like'' summary attention and full-layer integration.}
\label{fig:summary_attn_pattern}
\end{figure}

To understand \emph{why} hybrid architecture benefits long-context retrieval, we visualize the per-layer attention distributions of \textbf{KSA} and \textbf{Hybrid-KSA} on an out-of-window NIAH example in Figure~\ref{fig:summary_attn_pattern}.
We compare the first block, \textit{i.e.,} shallow layers L0-3 and the last block \textit{i.e.,} deep layers L24-27; within each row, the y-axis is shared across the two models so that attention magnitudes are directly comparable.

Two qualitative patterns emerge in \textbf{Hybrid-KSA} that are absent in \textbf{KSA}:

i) \textit{\textbf{Hybird SA} attends to summary tokens more frequently.}
In shallow SA layers (most clearly in L2), the hybrid model develops a periodic ``comb'' pattern that attends to past summary tokens at every chunk boundary, while \textbf{KSA} exhibits only a single weak peak close to the needle position.

ii) \textit{The interleaved full-attention layers (L3, L27, labeled ``[F]'') serve as cross-chunk integrators.}
Their attention maps show a sharp spike at the needle chunk while \textbf{KSA} remains flat, providing an explicit token-level retrieval indicator that \textbf{KSA} must instead approximate only summary tokens.

iii) \textit{Both models focus heavily on the earliest summary tokens in the deep block with notably different shapes.}
Hybrid-KSA's sink spreads across roughly six chunks, while the pure model's sink collapses onto only two, suggesting that the hybrid distributes its ``register'' role over a wider sink basin.

Taken together, \textit{the shallow comb pattern and the full-layer integrator provide two concerted retrieval indicators that \textbf{KSA} lacks}, which we hypothesize is the key reason \textbf{Hybrid-KSA} retrieves out-of-window needles more reliably thus possessing robust long-context capacity.

\section{Related Works}
In this section, we briefly review the latest efforts of long-context modeling.

\subsection{Efficient Attention Variants}

Multi-head attention (MHA)~\cite{vaswani2017attention} has become a cornerstone of modern deep learning, serving as a key building block across language~\cite{2018gpt1,devlin2019bert}, vision~\cite{dosovitskiy2020image,ramesh2021zero}, and video generation~\cite{peebles2023scalable}.
Despite its remarkable success, MHA faces two major bottlenecks when scaling to long contexts:
i) the quadratic cost of computing the full \(N \times N\) attention score matrix during training and long-sequence prefill; and
ii) the linearly growing KV cache during auto-regressive generation, which imposes substantial memory and bandwidth pressure.
To address these limitations, prior works have explored efficient attention mechanisms from several complementary directions.

The first line of work reduces the memory footprint of KV cache during decoding.
Multi-query attention (MQA)~\cite{shazeer2019mqa} lets all query heads share a single key-value head, while grouped-query attention (GQA)~\cite{ainslie2023gqa} generalizes this idea by allowing groups of query heads to share key-value heads.
Multi-head Latent Attention (MLA)~\cite{liu2024mla} further compresses KV states by projecting them into a low-dimensional latent space.
With matrix absorption during inference, MLA only stores compact latent vectors in the KV cache, substantially reducing memory overhead while preserving much of the expressiveness of MHA.
These methods are effective for memory-efficient decoding, but they do not directly eliminate the quadratic computation during long-context prefill, and they still rely on token-level cache to access historical context.

The second category reduces attention computation by limiting the attention scope.
\textbf{Local attention} mechanisms, such as sliding-window and sliding-chunk approaches~\cite{child2019generating,beltagy2020longformer,jiang2023diego,team2024gemma}, restrict each query to attend only to KV states within a local window of size \(W\), reducing the complexity from \(\mathcal{O}(N^2)\) to \(\mathcal{O}(WN)\).
These methods are simple and hardware-friendly, but weaken access to distant contexts that are crucial for long-context reasoning and retrieval.
\textbf{Sparse attention} variants~\cite{yang2026longlive,zhang2025spargeattn,yuan2025native,liu2025dsa} address this limitation by selecting a small subset of \(K\) important KV tokens or blocks according to heuristic scores, learned routing, or attention statistics, and computing attention only over the selected subset.
This reduces the attention cost to \(\mathcal{O}(KN)\) while potentially preserving selected long-range dependencies.
However, sparse attention typically still requires maintaining a large KV cache, and dynamic token or block selection introduces additional routing, indexing, and hardware-efficiency challenges.

The third direction replaces full softmax attention with linear, recurrent, or state-space sequence-mixing mechanisms.
Kernelized linear attention methods approximate or reformulate the softmax kernel with feature maps, avoiding explicit materialization of the full attention matrix.
For example, Retentive Networks~\cite{sun2023retentive} and Eagle~\cite{peng2024eagle}, maintain recurrent memory states with decay mechanisms, enabling efficient step-wise inference.
In parallel, state-space and gated recurrent models, such as Mamba~\cite{gu2024mamba} and Gated Delta Net~\cite{yang2025gated}, model long-range dependencies through selective fixed-dimensional states whose memory usage does not grow with the context length.
These methods offer attractive inference efficiency, but compressing the entire history into a fixed-size state can lose fine-grained token-level information, often leading to performance degradation on tasks that require precise long-context retrieval or reasoning.

Overall, existing efficient attention variants still face a fundamental performance-efficiency trade-off.
KV-cache compression reduces memory usage but does not fully address prefill cost; local and sparse attention reduce computation but may either lose distant information or retain large KV caches; and linear or recurrent variants improve efficiency but may sacrifice fine-grained retrieval ability.

\subsection{Hybrid Attention Architecture}

To better balance the strengths of full attention and efficient attention, recent research has increasingly explored hybrid attention architectures.
Rather than replacing all softmax attention layers with a single efficient mechanism, these models combine dense, sparse, local, recurrent, or state-space components within one architecture.
A widely adopted practice is to allocate a small fraction of full-attention layers among more efficient layers, thereby preserving global recall and expressiveness while reducing KV-cache cost and computation.

Hybrid-H3~\cite{fu2023hungry} builds a predominantly SSM-based backbone with only a small number of attention layers, showing that a limited dose of softmax attention can mitigate the expressivity gap of pure SSMs in language modeling.
RecurrentGemma~\cite{botev2024recurrentgemma} combines gated linear recurrences with sliding-window attention at a fixed ratio, leveraging the complementary benefits of recurrent memory and local sparse attention.
Mamba-2-Hybrid~\cite{waleffe2024empirical} constructs an 8B-parameter model from a mixture of Mamba-2 layers, full-attention layers, and MLP layers, demonstrating that a modest attention budget can support strong long-context reasoning.
Jamba~\cite{lieber2024jamba,team2024jamba} further scales the hybrid paradigm by interleaving Mamba blocks with Transformer and Mixture-of-Experts (MoE) layers, enabling long-context modeling with improved throughput and memory efficiency.

Beyond layer-level interleaving, several works introduce more structured hybridization strategies.
Zamba~\cite{glorioso2024zamba,glorioso2024zamba2} shares a global attention block across multiple Mamba blocks, obtaining benefits from global attention with improved parameter efficiency.
YOCO~\cite{sun2024you} separates sequence encoding and generation through a self-decoder and a cross-decoder: the self-decoder compresses the input sequence into compact KV states, while the cross-decoder repeatedly reuses them through cross-attention during generation, substantially reducing memory consumption across decoder layers.
Hymba~\cite{dong2025hymba} performs finer-grained hybridization by assigning different KV heads to softmax attention and SSM modules, allowing token-level information mixing across heterogeneous memory mechanisms.

These hybrid architectures reveal an important trend: different efficient attention mechanisms are often complementary rather than mutually exclusive.
Full attention provides high-fidelity global retrieval, local attention preserves nearby details with low cost, and recurrent or state-space modules offer compact memory for efficient generation.
However, most existing hybrid designs combine these components at the layer, block, or head level, and rely on carefully engineered architectural schedules to balance efficiency and expressiveness.
They usually lack an intrinsic interface that explicitly connects local fine-grained attention with compressed distant context.

In contrast, our proposed KSA introduces summary tokens as an explicit bridge between local information and long-range memory.
Instead of simply interleaving full and efficient layers, KSA performs hybridization inside the attention mechanism itself: local tokens preserve fine-grained short-range dependencies, while summary tokens provide compact access to distant contexts.
This intrinsic mixing mechanism enables KSA to reduce long-context attention and KV-cache overhead while maintaining strong modeling capacity for both local details and global information.

\subsection{Sequence-Level Token Compression}
\label{sec:sequence_level_token_compression}

Beyond optimizations along the layer or attention-scope dimensions, another line of work directly performs semantic-level compression along the sequence dimension by introducing a small number of learnable \emph{gist} or \emph{summary} tokens to condense historical contexts, so that distant information can be accessed through these compressed virtual tokens rather than full token-level history. Gisting~\cite{mu2023learning} introduces learnable gist tokens to compress an instruction-following prompt into a small set of cacheable activations via a modified attention mask, enabling efficient prompt reuse across inputs. AutoCompressors~\cite{chevalier2023adapting} extend this idea to long-context language modeling by recursively accumulating last-layer summary vectors as soft prompts across chunks, while Activation Beacon~\cite{zhang2025long} further pushes the compression into per-layer key-value activations and accumulates them across chunks during chunk-wise training. UniGist~\cite{deng2025unigist} removes the chunk-wise constraint with a unified sparse gist layout, in which every token can attend to all preceding gist tokens and a local raw window, enabling single-pass long-sequence training. More recent variants explore complementary directions on top of this line: DAST~\cite{chen-etal-2025-dast} dynamically reallocates a fixed soft-token budget across chunks based on information density, with the per-chunk allocation determined at inference time, while AdmTree~\cite{li2025admtree} organizes leaf gist tokens into an adaptive semantic tree and aggregates them with bidirectional self-attention to mitigate the unidirectional degradation of recursive compression.

Compared with the previous methods, we have the following design insights. \textbf{i) Hybrid attention architecture.} Instead of applying a single compressed pattern uniformly across all transformer layers, KSA interleaves summary-attention and full-attention layers at a 3:1 ratio, with the small dose of full-attention layers acting as a cross-chunk integrator (Sec.~\ref{sec:per_layer_attention_pattern_analysis}). \textbf{ii) Chunk-local non-recursive compression.} Each summary token attends strictly to the raw tokens within its own chunk, so every raw token is compressed exactly once and no recursive summary-of-summary aggregation is involved. \textbf{iii) Non-overlapping token routing.} Inside the local window a text token attends only to raw tokens, while distant context is exclusively accessed through summary tokens, eliminating double counting along the information path.

\section{Conclusion \& Future Works}
In this work, we revisit the long-context efficiency and highlight the sequence-level KV cache compression: rather than pursuing either strict linear KV cache reduction (GQA/MLA), we advocate \textit{sequence-level semantic compression with a moderate ratio $k$}, which maintains an $O(n/k)$ dependence on sequence length while preserving full addressability and interpretability over the distant history.
Specifically, we propose \textbf{Kwai Summary Attention (KSA)}, a drop-in efficient attention variant that i) periodically emits a learnable summary token for every text chunk, ii) routes local attention through a sliding chunk attention and distant attention through the accumulated summary tokens, and iii) guarantees a \textit{non-overlapping information conflict} so that every past chunk is either fully visible as raw text or accessed exclusively through its summary, avoiding both information loss and double counting.
Further, to support efficiency training and inference, we contribute: i) a block-sparse training and prefill kernel that exploits the structured sparsity of the KSA mask and avoids materializing the dense mask; ii) a contiguous, concatenation-free KV cache layout that encodes the visibility rule into memory addressing itself, such that every decoding step reads a single memory slice without gather or dynamic reshape; and ii) a hybrid-KSA architectural recipe that interleaves full-attention and KSA layers to balance retrieval fidelity with memory and compute cost.
On eleven benchmarks spanning long-context retrieval, knowledge, mathematics, and coding, our \textbf{Hybrid-KSA} configuration consistently establishes the smallest quality gap to full attention among all sub-quadratic alternatives, while surpassing full attention on the most context-intensive tasks such as RULER-128K.
These results support our motivation: sequence-level token compression provides  empirically favorable perspective for reducing the KV cache, and it is especially has the potential to support the complex agentic RL.

In the future, we will explore the following directions:
\textit{i) Sparse summary attention.} Our current design almost enabling the distant summary tokens are fully visible. Replacing this with a learned sparse retriever that selects summaries conditioned on the query may further improve long-context capability, particularly at extreme context lengths.
\textit{ii) KSA post-training.} We currently only perform the pre-training process, next we will exploring how KSA interacts with post-training, including supervised fine-tuning, preference optimization, reasoning reinforcement learning and multi-teacher OPD.
\textit{iii) Scaling laws of compression ratio.} The relationship between chunk size , model capacity, and task difficulty is not yet fully explored; establishing scaling laws for sequence-level compression, analogous to those for parameters and data, would guide a principle to scale next level intelligence.
\textit{iv) Unifying with OneRec/OneReason.} With the KV Cache friendly architecture, building a world knowledge enriched user KV cache memory is a promising iteration direction. Our next step is to build a unified generative recommendation foundation model on top of KSA, in which arbitrarily long user behavior trajectories are compressed into a hierarchy of summary tokens while recent interactions remain fully accessible through the local sliding window. We believe such an integration of KSA and OneRec/OneReason could bridge the gap between language-model-style world knowledge and recommendation-style user modeling, and serve as a foundation to build more smart recommendation system.

\printbibliography

\newpage
\quad \\
\quad \\

\section{Author List and Acknowledgement}

\noindent
\textbf{Core Contributors}\quad
Chenglong Chu,
Guorui Zhou\textsuperscript{*},
Guowang Zhang,
Han Li,
Hao Peng,
Hongtao Cheng\textsuperscript{*},
Hui Wang,
Jian Liang,
Jiangxia Cao\textsuperscript{$\dagger$},
Kun Gai,
Lingzhi Zhou,
Lu Ren,
Qi Zhang,
Ruiming Tang\textsuperscript{$\dagger$},
Ruitao Wang\textsuperscript{*},
Xinchen Luo\textsuperscript{*},
Yi Su\textsuperscript{*},
Zhiyuan Liang,
Ziqi Wang

\vspace{0.5em}
\noindent
\textbf{Contributors}\quad
Boyang Ding\textsuperscript{*},
Chengru Song,
Dunju Zang,
Jiao Ou,
Jiaxin Deng,
Jijun Shi,
Jinghao Zhang,
Junmin Chen,
Lejian Ren\textsuperscript{*},
Minxuan Lv,
Qianqian Wang,
Qigen Hu\textsuperscript{*},
Shiyao Wang\textsuperscript{*},
Siyang Mao,
Tao Wang,
Xingmei Wang,
Zhixin Ling,
Ziming Li,
Zixing Zhang\textsuperscript{*}

\vspace{0.5em}
\noindent
{\small All the authors listed alphabetically by first name.\quad \quad \textsuperscript{*}individuals who have departed from our team.\\
\textsuperscript{$\dagger$}Project Leaders.}

\end{document}